\documentclass[10pt,twocolumn,letterpaper]{article}

\usepackage[pagenumbers]{cvpr} 
\usepackage[accsupp]{axessibility}

\usepackage{array}
\usepackage{float}
\usepackage{arydshln}
\usepackage{multirow}

\usepackage[dvipsnames]{xcolor}

\def\paperID{10790} 
\def\confName{CVPR}
\def\confYear{2024}

\usepackage{color}
\usepackage{xcolor}
\usepackage[colorlinks=true,urlcolor=red]{hyperref}

\title{ArtAdapter: Text-to-Image Style Transfer using \\
Multi-Level Style Encoder and Explicit Adaptation}

\author{Dar-Yen Chen \quad Hamish Tennent \quad Ching-Wen Hsu\\
PicCollage, Taiwan\\
{\tt\small \{daryen.chen, hamish.tennent, gina.hsu\}@cardinalblue.com}\\
\small{\url{https://cardinalblue.github.io/artadapter.github.io/}}
}

\begin{document}
\twocolumn[{
\renewcommand\twocolumn[1][]{#1}%
\maketitle
\thispagestyle{empty}
\vspace{-9mm}
\begin{center}
    \centering
    \captionsetup{type=figure}
    \includegraphics[width=1\linewidth]{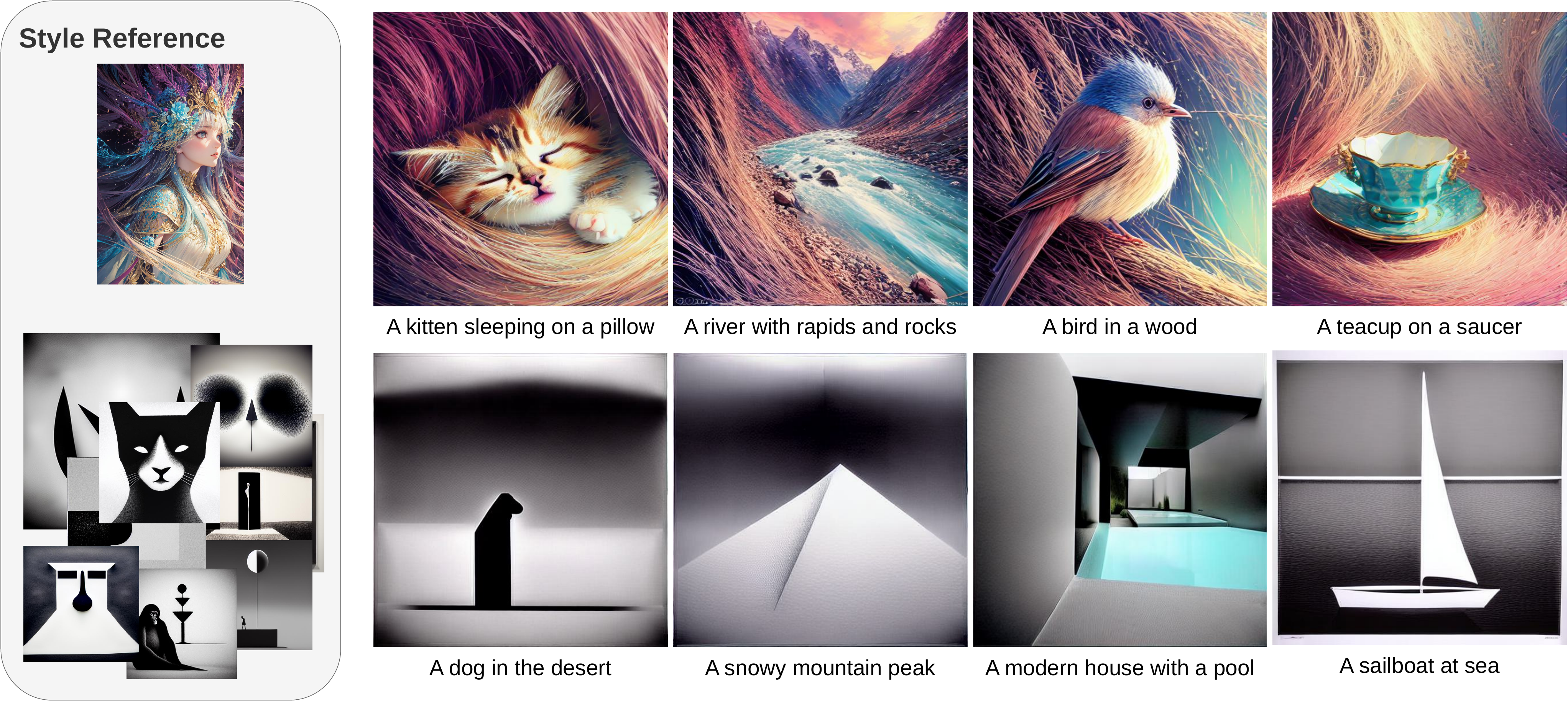}
    \vspace{-0.4cm}
    \captionof{figure}{Our framework is capable of capturing faithful style representation, from low-level delicate texture to high-level minimalism composition, in either single or multiple style references, closely adhering to the textual prompts.}
    \label{fig:banner}
    \vspace{0.2cm}
\end{center}
}]
\begin{abstract}
\vspace{-0.25cm}
This work introduces ArtAdapter, a transformative text-to-image (T2I) style transfer framework that transcends traditional limitations of color, brushstrokes, and object shape, capturing high-level style elements such as composition and distinctive artistic expression.
The integration of a multi-level style encoder with our proposed explicit adaptation mechanism enables ArtAdapter to achieve unprecedented fidelity in style transfer, ensuring close alignment with textual descriptions.
Additionally, the incorporation of an Auxiliary Content Adapter (ACA) effectively separates content from style, alleviating the borrowing of content from style references.
Moreover, our novel fast finetuning approach could further enhance zero-shot style representation while mitigating the risk of overfitting.
Comprehensive evaluations confirm that ArtAdapter surpasses current state-of-the-art methods.
\vspace{-0.4cm}
\end{abstract}    
\section{Introduction}
\label{sec:intro}

Bridging the realms of artificial intelligence and artistic creativity, text-to-image (T2I) style transfer \cite{Kwon_2022_CVPR, fu2022ldast, kwon2023diffusionbased} stands out as a captivating domain that masterfully transforms textual descriptions into visually rich and stylistic representations.
The core challenge lies in not just generating text-resonant images but infusing them with artistic depth and nuance, spanning from delicate brushstrokes to bold compositional elements—thereby capturing the essence of artistic vision.
Conventional arbitrary style transfer (AST) methods \cite{zhang2020cast, Deng_2022_CVPR, huang2017arbitrary, chen2021artistic, An_2021_CVPR, NEURIPS2021_df535469, Liu_2021_ICCV, BMVC2017_114} typically struggle beyond low-level features such as medium and colors, failing to grasp the more sophisticated realms of artistic expression.
Diffusion approaches \cite{kwon2023diffusionbased, Zhang_2023_CVPR}, including Textual Inversion \cite{gal2023an}, DreamBooth \cite{ruiz2022dreambooth}, and Low-Rank Adaptation (LoRA) \cite{db_lora, hu2022lora}, have shown potential in style representation.
Yet, these methods are hindered by laborious finetuning processes and a tendency towards overfitting, leading to outputs overly influenced by the style references' content at the expense of the textual context.
Especially when dealing with the unique nature of artworks, these methods face difficulties in style transfer from a single reference.

Addressing these challenges, we present ArtAdapter, an innovative T2I style transfer framework utilizing style embeddings derived from a multi-level style encoder.
These style embeddings, representing various layers of artistic attributes, then intricately interplay with textual context within the text encoder.
The style information is refined further through the novel Explicit Adaptation mechanism, situated within the cross-attention layers of the diffusion model.
Here, the Explicit Adaptation focuses explicitly on the style pathway, while leaving the text pathway frozen. This ensures that the artistic elements, from foundational textures to the distinctive expression of the artworks, are faithfully and precisely represented in the generated images without compromising the textual generalizability.
Moreover, our approach incorporates the Auxiliary Content Adapter (ACA) during the training phase.
The ACA plays a vital role by offering weak content guidance, aiding in the separation of content structure from style references.
This ensures that the final images are not overwhelmed by the content of the style references, maintaining a clear representation of style elements and the narrative intent of the text prompts.
Furthermore, we introduce a fast finetuning method, which further refines the model's ability to capture nuanced style details.
This finetuning is effective for both single- and multi-image style references, where it achieves detailed style representation with minimal steps, greatly reducing the time and computational resources typically required. This innovation addresses the challenge of overfitting and speeds up the adaptation process.
A notable feature of ArtAdapter is its adeptness at style mixing.
Leveraging the multi-level style encoder, ArtAdapter can blend styles from various references, extracting and combining distinct stylistic elements at different levels.
This enables the production of images that blend a diverse array of artistic influences, thereby offering remarkable flexibility and creativity to style transfer.

The principal contributions of our work are summarized as follows:
(i) A groundbreaking T2I style transfer model, ArtAdapter, that harnesses a multi-level style encoder and the Explicit Adaptation mechanism to capture diverse levels of style representation, and ensures a subtle balance between style and textual semantics;
(ii) The Auxiliary Content Adapter (ACA) separates content structure from style references,  addressing the issue of content dominance from style references.
(iii) A fast finetuning approach that enables rapid and effective style adaptation;
(iv) ArtAdapter allows for style mixing across different hierarchical levels, enriching the creative potential of T2I style transfer.

\section{Related Work}
\label{sec:related}

\subsection{Text-to-Image Synthesis}
Text-to-image (T2I) synthesis has evolved remarkably, with early methods primarily leveraging Generative Adversarial Networks (GANs) \cite{goodfellow2014generative}.
A paradigm shift was marked by the introduction of discrete Variational Autoencoders (VAEs) \cite{NEURIPS2019_5f8e2fa1, esser2021taming} and autoregressive transformers \cite{NIPS2017_3f5ee243}, leading to a more stable training process and high-quality image generation as demonstrated by Esser \etal \cite{esser2021taming} and the DALL-E \cite{pmlr-v139-ramesh21a}.
Subsequent advancements with diffusion models \cite{ho2020denoising} further refined the generation with a gradual denoising process, offering enhanced control and fidelity in outputs, as showcased by DALL-E2 \cite{ramesh2022hierarchical}, Imagen \cite{saharia2022photorealistic}, and Stable Diffusion \cite{rombach2022high}.
Recent developments \cite{xu2023prompt, zhao2023uni, ye2023ip-adapter} have extended these models' capabilities in various context, with innovations like ControlNet \cite{Zhang_2023_ICCV} and T2I adapters \cite{mou2023t2i} for spatial semantics, enriching the conditioning options well beyond textual inputs.

\subsection{T2I Personalization}
In T2I personalization, the central goal is to tailor pretrained models to a specific concept using a collection of reference images.
Innovations like Textual Inversion \cite{gal2023an} and DreamBooth \cite{ruiz2022dreambooth} pioneered this direction, learning text embeddings and optimizing diffusion backbone respectively. 
The quest for efficiency has led to the development of Low-Rank Adaptation (LoRA) \cite{hu2022lora, db_lora}, which reduces parameters by decomposing residual weights into two low-rank estimated matrices.
SVDiff \cite{Han_2023_ICCV} and Lightweight DreamBooth \cite{ruiz2023hyperdreambooth} further refine this process using Singular Value Decomposition (SVD) and orthogonal incomplete basis within LoRA weight-space respectively.
Perfusion \cite{tewel2023keylocked} incorporates Rank-One Model Editing (ROME) \cite{meng2022locating}, exemplifies targeted model edits aligned with conceptual directions.
Meanwhile, HyperDreamBooth \cite{ruiz2023hyperdreambooth} enables rapid adaptation to new concepts through hypernetwork-initialized rank-1 residuals.
Beyond finetuning, various approaches \cite{shi2023instantbooth, xiao2023fastcomposer} deploy task-specific components for zero-shot adaptation to novel concepts, including Taming Encoder \cite{jia2023taming} and IP-Adapter \cite{ye2023ip-adapter} for content semantics.
Interestingly, T2I style transfer aligns closely with style personalization, treating artistic style as a unique conceptual entity.
Our work contributes to this evolving landscape by developing a novel zero-shot T2I style transfer framework whose performance can be further improved through fast finetuning.

\subsection{Style Transfer}
Traditionally, Style Transfer \cite{An_2021_CVPR, NEURIPS2021_df535469, Liu_2021_ICCV} has depended on extracting content from target images and matching style via second-order statistics \cite{huang2017arbitrary, BMVC2017_114}.
However, the biases \cite{chen2021artistic, Deng_2022_CVPR} inherent in statistics cripple the representations.
IEST \cite{chen2021artistic} and CAST \cite{zhang2020cast}, address this issue by introducing contrastive learning.
The advent of vision transformers \cite{dosovitskiy2020vit} has facilitated the capture of long-range features, as seen in StyleFormer \etal \cite{Wu_2021_ICCV} and StyTr$^2$ \cite{Deng_2022_CVPR}.
ArtFusion \cite{chen2023artfusion} presents an integration of diffusion models \cite{ho2020denoising} into style transfer, bypassing reliance on statistical matching.
Following the booming of T2I models, InST \cite{Zhang_2023_CVPR} exemplifies the growing interest in T2I style transfer, with semantics provided by textual prompt, learning artistic features and guiding the generating with the CLIP image encoder \cite{radford2021learning}.
T2I-Style-Adapter \cite{mou2023t2i} aligns style representation by harnessing cross-attention layers to process CLIP \cite{radford2021learning} style image embeddings.
Meanwhile, FreeDoM \cite{Yu_2023_ICCV} employs time-dependent energy guidance, where the similarity between style references and generated samples is measured using the distance of the Gram matrix.
Nonetheless, T2I style transfer confronts challenges such as content from style references overshadowing the textual context \cite{wang2023styleadapter} and obvious artefacts.
Addressing this, our framework introduces an auxiliary component to foster content-style disentanglement, mitigating the influence of content semantics from style references.

\section{Approach}
\label{sec:approach}

\textbf{Preliminary on T2I Diffusion Models.}
A Text-to-Image (T2I) diffusion model comprises two foundation components: the diffusion backbone $\epsilon_\theta$ and the text encoder $c_\theta$.
The diffusion backbone meticulously manages a process of adding and removing noise. The inference process begins with a distribution of random noise, which is gradually denoised step by step according to the textual prompt to form a coherent image.
The textual prompt, denoted as $y$, first undergoes tokenization and index-based lookup, linking it to text embeddings $F_{txt}$, a sequence of vectors.
These embeddings are further refined by the text encoder, which contextualizes the information, producing enriched text embeddings $E_{txt}$ that encapsulate the textual description's meaning, intent, and subtleties.
Typically, T2I diffusion models employ cross-attention layers to access and utilize the semantic information contained in $E_{txt}$.
The T2I diffusion models' objective is defined by the equation \cite{dhariwal2021diffusion}:

\begin{equation}
\mathcal{L} = \mathbb{E}_{x, y, \epsilon, t}\Bigr[||\epsilon_\theta(x_t, t, c_\theta(F_{txt})) - \epsilon||^2_2\Bigr],
\end{equation}

\noindent where $x$ is the image, and $\epsilon$ represents the noise component in the corrupted $x_t$ at timestep $t$.
This equation aims to estimate the noise involved in the diffusion process, which is crucial to the gradual denoising process, aligning samples with the text description.

\begin{figure}[t]
\begin{center}
   \includegraphics[width=1.\linewidth]{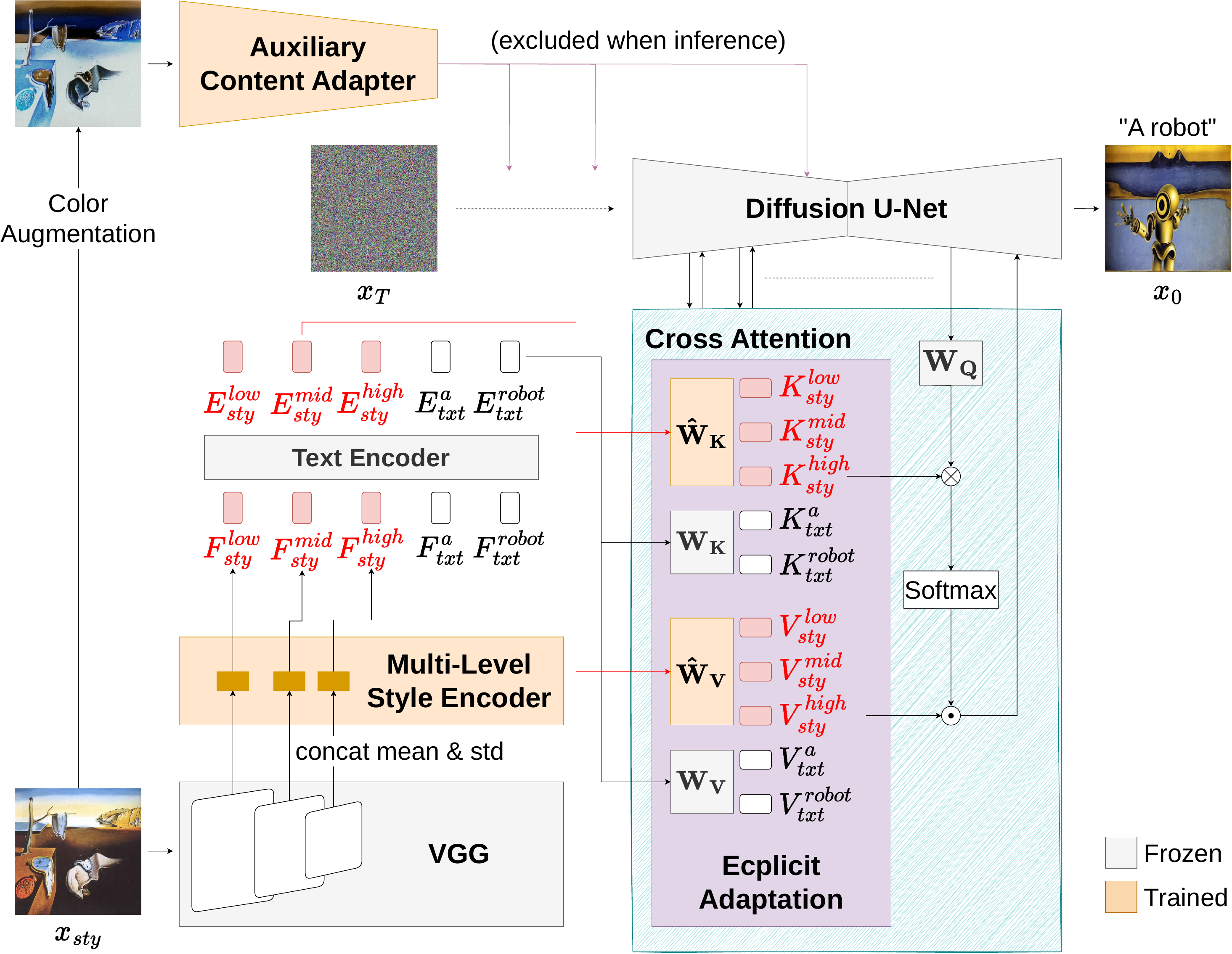}
\end{center}
   \vspace{-4mm}
   \caption{\textbf{Architecture of ArtAdapter.}
   Style embeddings, extracted through a cascade of a pretrained VGG \cite{7486599} followed by the multi-level style encoder, interact with text embeddings in the text encoder. In the cross-attention layers, the Explicit Adaptation exclusively optimizes the style-related projection to align outputs with style references. The Auxiliary Content Adapter provides weak content guidance during training, helping disentangle the content structure in the style reference. Our approach faithfully captures the style features without content semantics.
   }
   \vspace{-2mm}
\label{fig:architecture}
\end{figure}

\noindent\textbf{Overall Framework.}
In Figure \ref{fig:architecture}, we present the architecture of our ArtAdapter, utilizing a pretrained diffusion model as its backbone.
Our method instructs the diffusion model in articulating diverse styles through learnable style embeddings.
For encoding the style reference $x_{sty}$ into style embeddings $F_{sty}$, ArtAdapter employs a multi-level style encoder to process activations of pretrained VGG network \cite{7486599}.
Subsequently, these style embeddings are concatenated with text embeddings $F_{txt}$, and processed through the text encoder to obtain the style encodings $E_{sty}$.
To allow the diffusion backbone to adapt to these style concepts, we introduce the Explicit Adaptation mechanism within the cross-attention layers. This mechanism exclusively adapts to style encodings while preserving the integrity of text encoding. This ensures the robust generalization of the pretrained diffusion model while enabling precise adaptation to complex style nuances.
Furthermore, our framework incorporates the Auxiliary Content Adapter (ACA) - a key component during training, that is excluded when inference.
ACA provides rough content structure information to the UNet backbone \cite{ronneberger2015u}, helping eliminate the influence of the content semantics in the style reference.
Conclusively, our fast finetuning method is designed to capture more nuanced style characteristics, applicable to both individual and collective style references.

\subsection{Multi-Level Style Encoder}
In the realm of style transfer, second-order statistics of VGG \cite{7486599} activations, specifically the mean and standard deviation, have been crucial for their effectiveness in achieving style similarity \cite{Huang2017ArbitraryST, BMVC2017_114}.
These statistics lack overt content information, enabling more unadulterated style features.
A core of our ArtAdapter lies in the extraction of multi-level style features, amplifying the expressiveness and interpretability of the style transfer.
Specifically, we select VGG activations - \texttt{relu3\_3}, \texttt{relu4\_3}, and \texttt{relu5\_3} - to represent low-, mid-, and high-level features, respectively.
Once we concatenate these activations' mean and standard deviation \cite{huang2017arbitrary, chen2023artfusion}, low-, mid-, and high-leve style embeddings $F_{sty}^{low}$, $F_{sty}^{mid}$, $F_{sty}^{high}$ are extracted using distinct MLPs.
These style embeddings interact with the textual context in the text encoder and infuse the nuanced style information into the diffusion model. 

Intuitively, this mechanism is akin to captioning the style reference with pseudo-words, resulting in stylized prompts in the form of "$[style_{low}]$ $[style_{mid}]$ $[style_{high}]$ a robot", where $[style_{low/mid/high}]$ serve as learnable descriptors for styles at each level.
Notably, our implementation generates 3 embeddings per level, culminating in a comprehensive style embedding $F_{sty}$ of length 9.
These multi-level style embeddings provide rich style context, significantly enhancing the style fidelity.

\subsection{Explicit Adaptation}

Significant advancements in adapting pretrained diffusion models to new contexts, notably with efficient methods like LoRA \cite{hu2022lora}, have demonstrated effective adaptation with fewer parameters.
To enhance the integration of style within the diffusion backbone, we introduce Explicit Adaptation, a novel adaptation mechanism different from pursuing minimal parameters.
Within the cross-attention layers, the Explicit Adaptation mechanism is meticulously applied to the key and value projections of the style encodings $E_{sty}$, while leaving the pathways of the text embeddings $E_{txt}$ frozen.
The output $h$ of K- or V-projection in a cross-attention layer can be represented as:

\begin{equation}
h = W \{E_{sty}, E_{txt}\} + \alpha \Delta W \{E_{sty}, 0\}
\label{eq:explicit_adaptation}
\end{equation}

\noindent where $W$ represents the original weights, which are frozen to maintain existing knowledge, $\alpha$ is a learnable scale, and $\Delta W$ is the residual weight focusing on the style encodings.

The term "Explicit" signifies a direct and intentional approach, not only in the operational dynamics of the adaptation but also in the model's learning objectives.
This deliberate demarcation ensures the adaptation process is sharply concentrated on the incorporation of style nuances hidden in the style encodings.
This innovation enables our ArtAdapter to refine style representations without compromising the established robust linguistic knowledge base.
The merits of Explicit Adaptation, particularly in its role in the precise acquisition of fine-grained style features, will be further dissected and validated in Section \ref{sec:ablation}.

\subsection{Auxiliary Content Adapter}
One of the major challenges in T2I style transfer is the unintentional overpowering of the style reference's content within the generated images, eclipsing the textual semantics.
Even conditioning on VGG \cite{7486599} statistics features without content structure, due to certain highly recurrent patterns in the dataset, like human faces \cite{chen2023artfusion}, the model might become prone to overfitting specific statistical features of content, leading it to transfer content patterns from the style reference.
To address this challenge, we introduce the Auxiliary Content Adapter (ACA) into our framework, a variant of the T2I adapter \cite{mou2023t2i}.
It takes the image $x$ as input, providing the diffusion backbone with essential content cues.
This ensures that the style components in ArtAdapter, including the multi-level style adapter and the Explicit Adaptation, concentrate solely on capturing pure style features, avoiding any unwanted diversion towards mimicking content semantics from the style reference.

To calibrate the influence of ACA, its features are infused within the deepest input block of the UNet backbone \cite{ronneberger2015u}, and positioned to impact only the initial denoising stages (20\% in this work), where the model delineates only the rudimentary content outlines.
Moreover, we subject the ACA's input to a series of random color augmentation, including color jitter, greyscale, inversion, and polarization, to prevent the model from learning the style's color palette using ACA — reserving color learning for the style encoder.
These approaches confine the ACA's function to offering only the rough content structure, devoid of intricate details, thus paving the way for a more nuanced style representation.

It's imperative to underscore that ACA's role is confined to the training phase. In an ideal training process, the ACA and textual prompts, provide the content structure and semantics, respectively.
During inference, ACA and style references' captions are excluded, allowing our ArtAdapter to draw exclusively from the learned style features. This ensures that the generated images present only the desired style characteristics without content from style references.

\begin{figure*}[t]
\begin{center}
   \includegraphics[width=1.\linewidth]{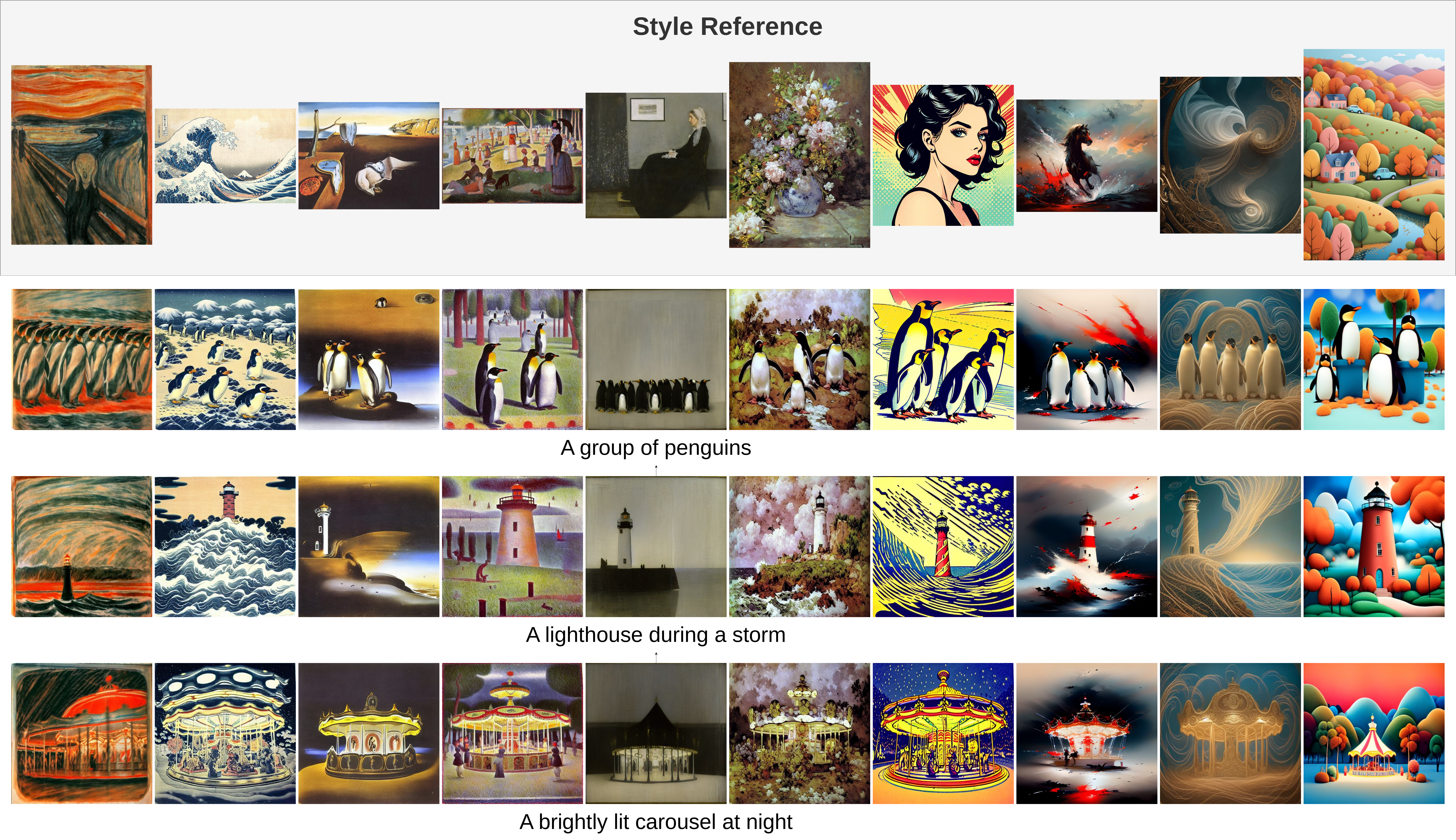}
\end{center}
   \vspace{-2mm}
   \caption{\textbf{Qualitative results.} This collection of images exhibits the ArtAdapter's capability to present faithful style representation across diverse artworks without compromising on semantics, showcasing the versatility and deep understanding of artistic and textual contexts.}
\label{fig:showcase}
\end{figure*}

\subsection{Fast Finetuning and Multiple Style References}
Building on the Explicit Adaptation mechanism, we introduce a fast finetuning method to enhance the precision and subtlety of zero-shot style representation.
The crux of this finetuning lies in the following equation:

\vspace{-3mm}
\begin{equation}
h = W \{E_{sty}, E_{txt}\} + \alpha \Delta W \{E_{sty}, 0\} + \{\Delta h_{sty}, 0\}
\label{eq:explicit_finetune}
\end{equation}

\noindent$\Delta h_{sty}$ is the residual vector that directly adjusts the style-related activations, refining the model's ability to mirror the characteristics of the style references.
During finetuning, all components besides the vector residuals remain frozen, ensuring an efficient adaptation process focused on the enhancement of style representation.
The principle of explicit adaptation mechanism ensures that the textual context pathways remain intact, offering a safeguard against overfitting—a common challenge in single-image personalization.
This finetuning is characterized by its efficiency. With only minimal extra parameters from the rank-1 residual vector, and a focus on style operations, the process is expedited to dozens of steps, typically completed within minutes.
Notably, this approach can be easily extended to multi-reference style transfer. By averaging the style embeddings from all references, the model can optimize the vector residuals to reflect the aggregate style.

\subsection{Style Mixing}

Our innovative style mixing approach leverages the multi-level style encoder to expand the T2I style transfer's horizons.
With a trio of style reference images, we extract their respective style embeddings at low, mid, and high levels. These are subsequently constructed into compositional style embeddings by concatenation.
Such a fusion of embeddings enables ArtAdapter to synthesize a new, mixed style that simultaneously exhibits features from the individual styles at corresponding hierarchical levels.
Furthermore, this method of style mixing can easily extend to styles that have undergone finetuning, by applying the residual vectors to the corresponding entries.

\section{Experiments}
\label{sec:experiments}

\noindent\textbf{Data.}
For training, we harness the LAION AESTHETICS \cite{schuhmann2022laionb} and WikiArt \cite{phillips:2011:wiki} datasets. The WikiArt dataset has been captioned using BLIP-2 \cite{li2023blip}. For testing, we adopt a test dataset comprising 35 prompts, including some derived from Wang \etal \cite{wang2023styleadapter}. We collect 10 styles for both single- and multi-reference style transfer separately, culminating in a total of 700 image pairs for the test dataset.

\noindent\textbf{Evaluation Metrics.}
We employ the CLIP \cite{radford2021learning} metrics for objective assessment, including similarity for text and style.
The aesthetics score \cite{aestheticscore} functions as an indicator, gauging the aesthetic appeal of the generated images.
Additionally, we undertake a user study involving 212 participants utilizing a 7-point scale, with 5 tuples per user, to assess subjective text and style accuracy, alongside the overall quality.

\noindent\textbf{Implementation Details.}
This work leverages the SD V1.5 \cite{rombach2022high} as the backbone. We employ an AdamW optimizer \cite{loshchilov2018decoupled} with a batch size of 16, a learning rate of 1e-4 for the multi-level style encoder and the ACA, and a reduced learning rate of 1e-7 for Explicit Adaptation residual weights. Specifically, we apply LoRA \cite{db_lora, hu2022lora} to reduce the parameters. During the finetuning, we optimize the vector residual using a learning rate of 0.02 over 25 iterations, taking $\sim \!\!1$ minute. All experiments are undertaken on a single Nvidia GTX 4090 GPU. For inference, we set the sampling \cite{song2021denoising} steps to 50, with the classifier-free guidance scale \cite{ho2021classifierfree} being fixed at 9.
Unless otherwise specified, all reported results were obtained with finetuning.

\begin{figure}[t]
\begin{center}
   \includegraphics[width=1.\linewidth]{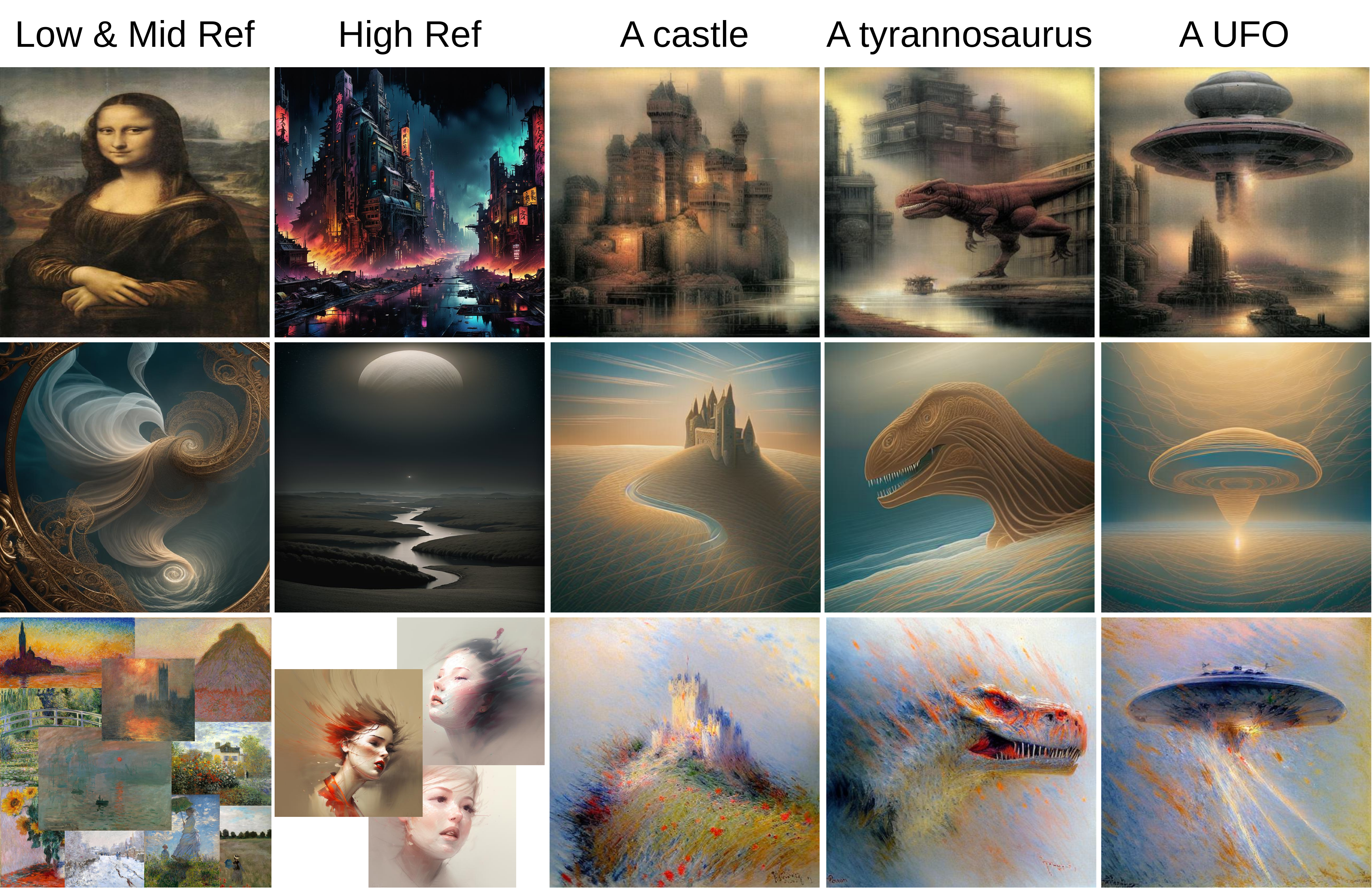}
\end{center}
   \vspace{-4mm}
   \caption{\textbf{Illustration of Style Mixing.} The seamless integrations of the two styles reflect the distinct contributions of style features from different hierarchical levels to the final images and demonstrate the remarkable flexibility of ArtAdapter.}
   \vspace{-2mm}
\label{fig:mixing}
\end{figure}

\begin{figure*}[t]
\begin{center}
   \includegraphics[width=1.\linewidth]{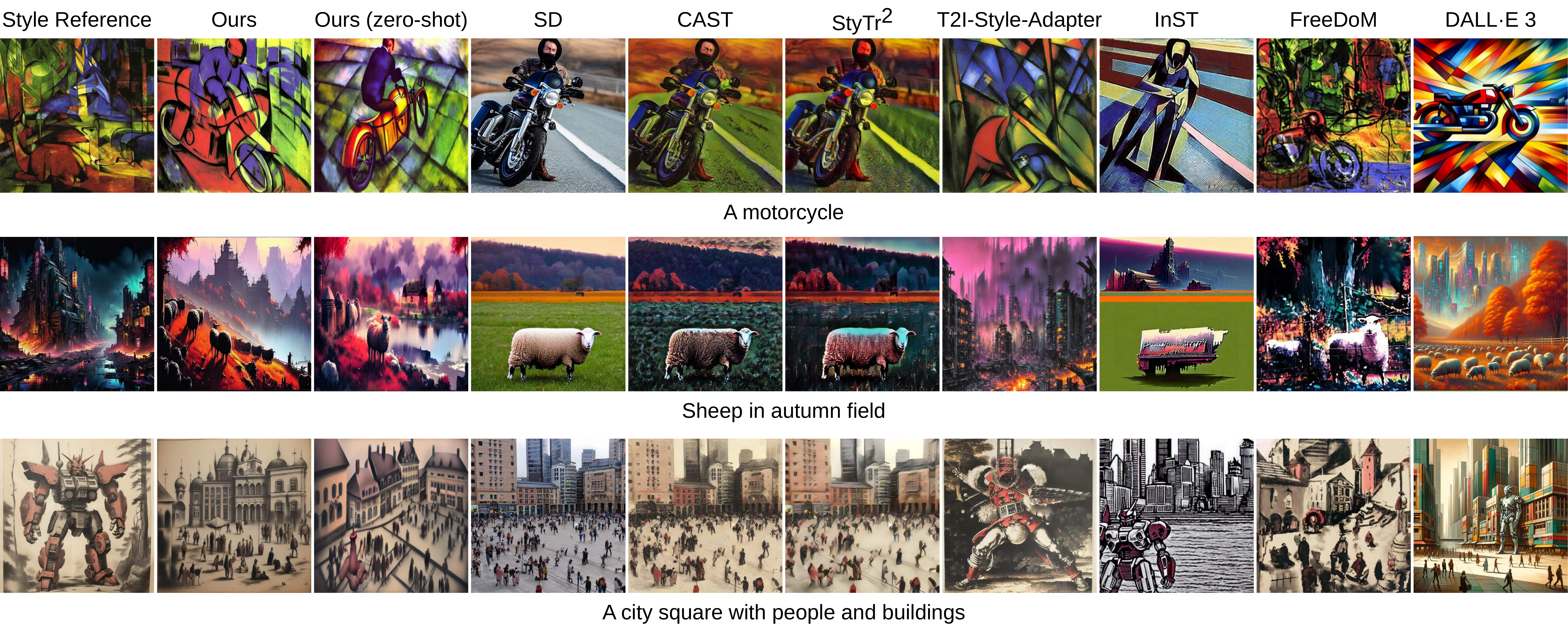}
\end{center}
   \vspace{-4mm}
   \caption{\textbf{Qualitative comparison on single style reference.} Our results showcase ArtAdapter's superior style alignment over other approaches \cite{zhang2020cast, Deng_2022_CVPR, Zhang_2023_CVPR, betker2023improving}. Note that the SD \cite{rombach2022high} column works as content target images for conventional AST models \cite{zhang2020cast, Deng_2022_CVPR}.}
   \vspace{-1mm}
\label{fig:single}
\end{figure*}

\begin{figure*}[t]
\begin{center}
   \includegraphics[width=1.\linewidth]{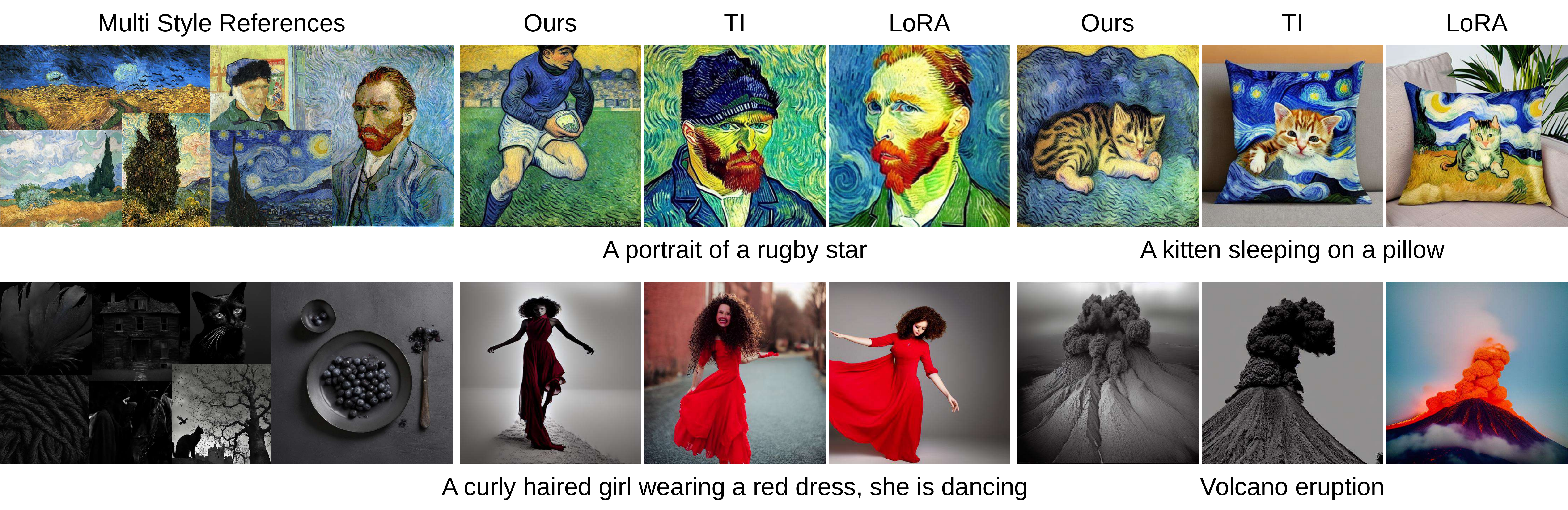}
\end{center}
   \vspace{-4mm}
   \caption{\textbf{Qualitative comparison on multiple style references.} Compared to TI \cite{gal2023an} and LoRA \cite{db_lora, hu2022lora}, our ArtAdapter provides greater consistency and coherence, while preventing content borrowed from the style references to the results.}
   \vspace{-2mm}
\label{fig:multi}
\end{figure*}

\subsection{Qualitative Evaluation}

The ArtAdapter's prowess is showcased in Figure \ref{fig:banner}, where it captures nuanced styles from various either single or multiple style references, while faithfully reflecting the textual prompts.
This is particularly evident in the top row of Figure \ref{fig:banner}, where a clear facial structure is depicted in the style reference, a challenging scenario often encountered in T2I style transfer.
Remarkably, our model adeptly reproduces the ornate texture and aesthetic of the style reference without erroneously incorporating the facial structure into the final results.
The bottom row of Figure \ref{fig:banner} illustrates ArtAdapter's capability to go beyond mere low-level style representation.
Here, we observe not only the replication of black-and-white tones and geometric patterns but, more crucially, the high-level abstract composition that resonates with the style reference collection.
This demonstrates the multi-level style encoder and Explicit Adaptation mechanism's collaborative efficacy in rendering styles that are both authentic and expressive.

Figure \ref{fig:showcase} provides further insights into ArtAdapter's capacity, showcasing side-by-side comparisons of various styles within the same textual context.
From the precise transfer of textures and tones to the emulation of geometric shapes and the overall composition, ArtAdapter showcases a deep understanding of the distinctive artistic concepts inherent in the style references.
ArtAdapter effectively circumvents borrowing content from style references, creatively reinterpreting artistic elements to yield visually compelling art that remains faithful to the textual descriptions.
This attests to the efficacy of the explicit mechanisms in our approach, which align the subject matter with the style, while preventing overfitting that commonly happens in single-image personalization finetuning.

In summary, our method facilitates T2I style transfer, advancing toward a more genuine and precise representation of styles. It encapsulates the essence of the artistic intent, offering a substantial leap.

\begin{table}[t]
\small
\begin{center}
    \begin{tabular}{llccc}
        \toprule
        Type & Approach               & Text $\uparrow$   & Style $\uparrow$  & Aesth $\uparrow$  \\
        \midrule
        \multirow{5}{*}{\parbox{0.8cm}{single\\ref}}   & CAST                  & 0.297 & 0.583 & 5.410 \\
                                        & StyTr$^2$  & 0.299 & 0.577 & 5.308 \\
                                        & T2I-Style-Adapter  & 0.181 & 0.822 & 5.705 \\
                                        & InST  & 0.235 & 0.660 & 5.255 \\
                                        & FreeDoM & 0.262 & 0.634 & 5.059 \\
                                        & Ours (zero-shot)   & 0.269 & 0.656 & 5.532 \\
                                        & \textbf{Ours} & \textbf{0.255} & \textbf{0.707} & \textbf{5.601} \\
        \midrule
        \multirow{3}{*}{\parbox{0.8cm}{multi\\ref}}   & TI                  & 0.265 & 0.678 & 5.712 \\
                                        & LoRA  & 0.254 & 0.678 & 5.701 \\
                                        & \textbf{Ours}  & \textbf{0.258} & \textbf{0.680} & \textbf{5.610} \\
        
        \bottomrule
    \end{tabular}
\end{center}
    \vspace{-3mm}
    \caption{\textbf{Quantitative Comparison.} ArtAdapter demonstrates superior balances between text and style similarity, as well as competitive aesthetics score, even with zero-shot.}
\label{tab:quant}
\end{table}

\begin{table}[t]
\small
\begin{center}
    \begin{tabular}{lccc}
        \toprule
        Model               & Text $\uparrow$   & Style $\uparrow$  & Quality $\uparrow$  \\
        \midrule
        TI                  & 4.49 & 4.35 & 4.38  \\
        LoRA  & 3.91 & 4.08 & 4.18 \\
        \textbf{Ours}  & \textbf{4.74} & \textbf{4.76} & \textbf{4.43} \\
        \bottomrule
    \end{tabular}
\end{center}
    \vspace{-3mm}
    \caption{\textbf{Perceptual Comparison.} ArtAdapter exhibits preeminence in all subjective metrics over TI \cite{gal2023an} and LoRA \cite{db_lora, hu2022lora}, highlighting its effectiveness in T2I multi-reference style transfer based on user preferences.}
    \vspace{-2mm}
\label{tab:user_study}
\end{table}

\subsection{Style Mixing}
\label{sec:mixing}

Figure \ref{fig:mixing} demonstrates ArtAdapter's proficiency in seamlessly mixing two styles within one image.
Achieving this fusion involves applying distinct styles to different hierarchical levels: one affects low- and mid-level features, while the other shapes high-level attributes.
This mixing reveals the significant roles of different levels in shaping the final image. The low and mid-levels primarily dictate tone and texture, as evidenced by the haziness of the "Mona Lisa" and Monet's brushstroke texture.
Conversely, high-level features dictate the overall compositional structure and artistic expression, as exemplified by the simplistic style in the second row.
This multi-level approach underscores our model's flexibility and interpretability in style transfer, facilitating a creative combination of artistic elements.

\subsection{Comparison with State-of-the-art Methods}
\label{sec:compare}

\noindent\textbf{Single Style Reference.}
In Figure \ref{fig:single}, our ArtAdapter is compared with conventional AST models, such as CAST \cite{zhang2020cast} and StyTr$^2$ \cite{Deng_2022_CVPR}, the diffusion-based T2I-Style-Adapter \cite{mou2023t2i}, InST \cite{Zhang_2023_CVPR}, FreeDoM \cite{Yu_2023_ICCV} and the advanced DALL-E3 \cite{betker2023improving} in single-reference T2I style transfer.
Utilizing SD \cite{rombach2022high} as for T2I generation, stylized by AST models, we have the T2I style transfer results of conventional AST approaches \cite{zhang2020cast, Deng_2022_CVPR}.
These models typically struggle to transcend basic color transfer, struggling with the conveyance of high-level style attributes, particularly abstract geometric configurations, and holistic style elements.
As diffusion-based approaches, T2I-Style-Adapter \cite{mou2023t2i}, while adept at representing the style, often discards the textual context, resulting in a loss of controllability.
InST \cite{Zhang_2023_CVPR} tends to produce results with unnatural style representation and unintended content borrowings from style references; 
FreeDoM \cite{Yu_2023_ICCV}, which relies on energy guidance, often introduces noticeable artefacts into the stylized images.
DALL-E3 \cite{betker2023improving}, leveraging ChatGPT \cite{chatgpt} to generate style descriptions, struggles in the actual reproduction of style features, revealing a bottleneck in prompt engineering.

Distinctly, our zero-shot ArtAdapter captures the original style's brushwork, texture, and overall aesthetic with exceptional fidelity, as evidenced by the quantitative results in Table \ref{tab:quant}.
This fidelity is further enhanced by the finetuning, leading to an alignment closely mirroring the original artworks.
The quantitative analysis, as shown in Table \ref{tab:quant},  highlights our style similarity score of 0.707 and an aesthetics score of 5.601, surpassing the benchmarks, except for T2I-Style-Adapter. While T2I-Style-Adapter edges out in style and aesthetics, the disproportionately low text similarity reveals a significant shortcoming in its T2I style transfer capability.  In contrast, our ArtAdapter excels, maintaining high style and aesthetics scores with only a minimal trade-off in text fidelity, which is 0.255.

\noindent\textbf{Multiple Style Reference.}
In assessing multi-reference T2I style transfer, our ArtAdapter is juxtaposed with Textual Inversion (TI) \cite{gal2023an} and Low-Rank Adaptation (LoRA) \cite{hu2022lora}, detailed in Figure \ref{fig:multi} and Table \ref{tab:quant}.
Despite the successes of TI \cite{gal2023an} and LoRA \cite{hu2022lora} in capturing subtle style nuances, they exhibit overfitting, as evidenced by their dominance of style reference content, such as the face of Van Gogh, leading to discrepancies with the textual prompts.
Instances such as the top-right corner of the sample manifest the incoherence in their style transfer, rooted in localized stylization problems.
Moreover, varying degrees of style assimilation within a single style reference in the bottom row reveals a lack of consistency.
These highlight the instability of TI \cite{gal2023an} and LoRA \cite{hu2022lora} in style transfer.
Our ArtAdapter excels in rendering outputs that accurately embody style features from the collection and eliminate content borrowing, with consistent representation in diverse textual contexts.
It nicely preserves the inherently strong generality of SD \cite{rombach2022high}.
Quantitative analysis demonstrates parity in performance among the methods, highlighting our approach's efficiency with a fast finetuning process of $\sim \!\!1$ minute.

\noindent\textbf{User Study.}
To evaluate the efficacy of our multi-reference T2I style transfer approach, a comprehensive user study, detailed in Table \ref{tab:user_study}, is conducted.
The study unveils a marked discrepancy in text and style precision, with ArtAdapter achieving 4.74 and 4.76 scores, respectively, thus underscoring its enhanced adherence to textual prompts and style references, surpassing TI \cite{gal2023an} and LoRA \cite{hu2022lora} benchmarks.
Moreover, the elevated quality score of 4.43 underscores the visual appeal of our generated images. 
Collectively, these results indicate that our approach provides an improved T2I style transfer experience from a user perspective.

\begin{figure}[t]
\begin{center}
    \includegraphics[width=1.\linewidth]{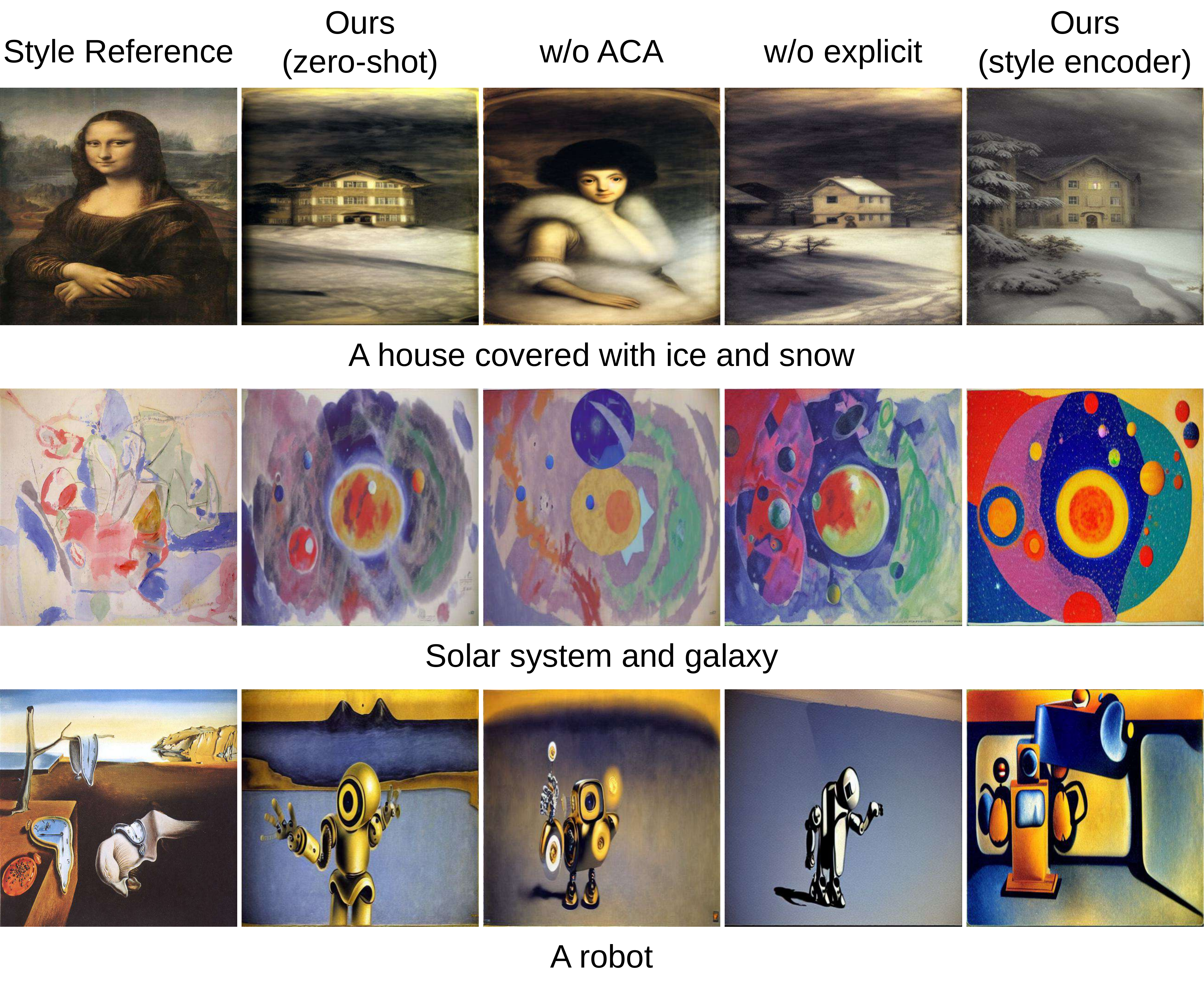}
\end{center}
    \vspace{-4mm}
    \caption{\textbf{Ablation research.} These comparisons highlight the importance of each mechanism in our framework.}
    \vspace{-2mm}
\label{fig:ablation}
\end{figure}

\subsection{Ablation Study}
\label{sec:ablation}

The ablative analysis illustrated in Figure \ref{fig:ablation} dissects our ArtAdapter, emphasizing the distinct contributions and impact of each core mechanism on style transfer performance.

\noindent\textbf{Auxiliary Content Adapter.}
The first row underscores the crucial role of the Auxiliary Content Adapter (ACA). In its absence, the model inappropriately adopts content semantics. The outputs will be dominated by content features in the style reference—such as the face in "Mona Lisa".
The ACA's function to separate content and style learning is confirmed; it empowers the model to preserve style features while filtering out content during inference, guaranteeing that the generated images adhere to the textual prompts.

\noindent\textbf{Explicit Adaptation.}
In the second and third rows, we scrutinize the effectiveness of the Explicit Adaptation mechanism. Without this mechanism, the model shows a marked decrease in its ability to accurately represent style across different levels.
This is evidenced by distorted colors, imprecise geometric formations, and a lack of depth in capturing the compositional elements and artistic intent, such as those found in Salvador Dali's surrealist works.
The explicit adaptation mechanism is thereby shown to be indispensable for the model's ability to internalize style contexts, generating outputs with high style fidelity.

\noindent\textbf{Exclusive Style Encoder.}
Sole reliance on the multi-level style encoder, without incorporating adaptation in the backbone, leads to significant degradation in style representation, especially obvious in the second and third rows.
The style encoder can only capture features such as imprecise color and rough texture patterns, failing to grasp the more profound stylistic expressions.
This clearly highlights the critical role of the diffusion backbone in adapting to various style contexts, affirming that the integration of all components is critical for optimal style transfer performance.

\section{Conclusion}
\label{sec:conclusion}

In this work, our innovative T2I style transfer framework, ArtAdapter, has effectively demonstrated its ability to synthesize images that faithfully align with given textual prompts and style references.
By employing a multi-level style encoder for nuanced style capture, the Explicit Adaptation for effective style integration, and the Auxiliary Content Adapter (ACA) for content-style separating, our ArtAdapter sets a new standard in achieving both style and text fidelity.
The implementation of our fast finetuning strategy significantly enhances the efficiency of style alignment, establishing our model as a benchmark in the realm of T2I style transfer.
However, ArtAdapter does have limitations, particularly in style mixing.
We observe that high-level style embeddings often inadvertently incorporate elements from lower levels, causing interference in the style mixing process.
We aim to refine the disentanglement of hierarchical style features to improve the authenticity and precision of style mixing in future works.

{
    \small
    \bibliographystyle{ieeenat_fullname}
    \bibliography{main}

\begin{thebibliography}{56}
\providecommand{\natexlab}[1]{#1}
\providecommand{\url}[1]{\texttt{#1}}
\expandafter\ifx\csname urlstyle\endcsname\relax
  \providecommand{\doi}[1]{doi: #1}\else
  \providecommand{\doi}{doi: \begingroup \urlstyle{rm}\Url}\fi

\bibitem[aes(2022)]{aestheticscore}
Clip+mlp aesthetic score predictor.
\newblock \url{https://github.com/christophschuhmann/improved-aesthetic-predictor}, 2022.

\bibitem[db_(2022)]{db_lora}
Low-rank adaptation for fast text-to-image diffusion fine-tuning.
\newblock \url{https://github.com/cloneofsimo/lora}, 2022.

\bibitem[cha(2023)]{chatgpt}
Chatgpt.
\newblock \url{https://openai.com/blog/chatgpt}, 2023.

\bibitem[An et~al.(2021)An, Huang, Song, Dou, Liu, and Luo]{An_2021_CVPR}
Jie An, Siyu Huang, Yibing Song, Dejing Dou, Wei Liu, and Jiebo Luo.
\newblock Artflow: Unbiased image style transfer via reversible neural flows.
\newblock In \emph{Proceedings of the IEEE/CVF Conference on Computer Vision and Pattern Recognition (CVPR)}, 2021.

\bibitem[Betker et~al.(2023)Betker, Goh, Jing, Brooks, Wang, Li, Ouyang, Zhuang, Lee, Guo, Manassra, Dhariwal, Chu, Jiao, and Ramesh]{betker2023improving}
James Betker, Gabriel Goh, Li Jing, Tim Brooks, Jianfeng Wang, Linjie Li, Long Ouyang, Juntang Zhuang, Joyce Lee, Yufei Guo, Wesam Manassra, Prafulla Dhariwal, Casey Chu, Yunxin Jiao, and Aditya Ramesh.
\newblock Improving image generation with better captions.
\newblock 2023.

\bibitem[Chen(2023)]{chen2023artfusion}
Dar-Yen Chen.
\newblock Artfusion: Controllable arbitrary style transfer using dual conditional latent diffusion models, 2023.

\bibitem[Chen et~al.(2021{\natexlab{a}})Chen, Zhao, Wang, Ming, Zuo, Li, Xing, and Lu]{chen2021artistic}
Haibo Chen, Lei Zhao, Zhizhong Wang, Zhang~Hui Ming, Zhiwen Zuo, Ailin Li, Wei Xing, and Dongming Lu.
\newblock Artistic style transfer with internal-external learning and contrastive learning.
\newblock In \emph{Advances in Neural Information Processing Systems}, 2021{\natexlab{a}}.

\bibitem[Chen et~al.(2021{\natexlab{b}})Chen, zhao, Wang, Zhang, Zuo, Li, Xing, and Lu]{NEURIPS2021_df535469}
Haibo Chen, lei zhao, Zhizhong Wang, Huiming Zhang, Zhiwen Zuo, Ailin Li, Wei Xing, and Dongming Lu.
\newblock Artistic style transfer with internal-external learning and contrastive learning.
\newblock In \emph{Advances in Neural Information Processing Systems}, 2021{\natexlab{b}}.

\bibitem[Deng et~al.(2022)Deng, Tang, Dong, Ma, Pan, Wang, and Xu]{Deng_2022_CVPR}
Yingying Deng, Fan Tang, Weiming Dong, Chongyang Ma, Xingjia Pan, Lei Wang, and Changsheng Xu.
\newblock Stytr2: Image style transfer with transformers.
\newblock In \emph{Proceedings of the IEEE/CVF Conference on Computer Vision and Pattern Recognition (CVPR)}, 2022.

\bibitem[Dhariwal and Nichol(2021)]{dhariwal2021diffusion}
Prafulla Dhariwal and Alexander~Quinn Nichol.
\newblock Diffusion models beat {GAN}s on image synthesis.
\newblock In \emph{Advances in Neural Information Processing Systems}, 2021.

\bibitem[Dosovitskiy et~al.(2021)Dosovitskiy, Beyer, Kolesnikov, Weissenborn, Zhai, Unterthiner, Dehghani, Minderer, Heigold, Gelly, Uszkoreit, and Houlsby]{dosovitskiy2020vit}
Alexey Dosovitskiy, Lucas Beyer, Alexander Kolesnikov, Dirk Weissenborn, Xiaohua Zhai, Thomas Unterthiner, Mostafa Dehghani, Matthias Minderer, Georg Heigold, Sylvain Gelly, Jakob Uszkoreit, and Neil Houlsby.
\newblock An image is worth 16x16 words: Transformers for image recognition at scale.
\newblock 2021.

\bibitem[Esser et~al.(2021)Esser, Rombach, and Ommer]{esser2021taming}
Patrick Esser, Robin Rombach, and Bjorn Ommer.
\newblock Taming transformers for high-resolution image synthesis.
\newblock In \emph{Proceedings of the IEEE/CVF conference on computer vision and pattern recognition}, 2021.

\bibitem[Fu et~al.(2022)Fu, Wang, and Wang]{fu2022ldast}
Tsu-Jui Fu, Xin~Eric Wang, and William~Yang Wang.
\newblock {Language-Driven Artistic Style Transfer}.
\newblock In \emph{European Conference on Computer Vision (ECCV)}, 2022.

\bibitem[Gal et~al.(2023)Gal, Alaluf, Atzmon, Patashnik, Bermano, Chechik, and Cohen-or]{gal2023an}
Rinon Gal, Yuval Alaluf, Yuval Atzmon, Or Patashnik, Amit~Haim Bermano, Gal Chechik, and Daniel Cohen-or.
\newblock An image is worth one word: Personalizing text-to-image generation using textual inversion.
\newblock In \emph{The Eleventh International Conference on Learning Representations}, 2023.

\bibitem[Ghiasi et~al.(2017)Ghiasi, Lee, Kudlur, Dumoulin, and Shlens]{BMVC2017_114}
Golnaz Ghiasi, Honglak Lee, Manjunath Kudlur, Vincent Dumoulin, and Jonathon Shlens.
\newblock Exploring the structure of a real-time, arbitrary neural artistic stylization network.
\newblock In \emph{BMVC}, 2017.

\bibitem[Goodfellow et~al.(2014)Goodfellow, Pouget-Abadie, Mirza, Xu, Warde-Farley, Ozair, Courville, and Bengio]{goodfellow2014generative}
Ian~J Goodfellow, Jean Pouget-Abadie, Mehdi Mirza, Bing Xu, David Warde-Farley, Sherjil Ozair, Aaron Courville, and Yoshua Bengio.
\newblock Generative adversarial networks.
\newblock In \emph{Proc.\ NeurIPS}, 2014.

\bibitem[Han et~al.(2023)Han, Li, Zhang, Milanfar, Metaxas, and Yang]{Han_2023_ICCV}
Ligong Han, Yinxiao Li, Han Zhang, Peyman Milanfar, Dimitris Metaxas, and Feng Yang.
\newblock Svdiff: Compact parameter space for diffusion fine-tuning.
\newblock In \emph{Proceedings of the IEEE/CVF International Conference on Computer Vision (ICCV)}, 2023.

\bibitem[Ho and Salimans(2021)]{ho2021classifierfree}
Jonathan Ho and Tim Salimans.
\newblock Classifier-free diffusion guidance.
\newblock In \emph{NeurIPS 2021 Workshop on Deep Generative Models and Downstream Applications}, 2021.

\bibitem[Ho et~al.(2020)Ho, Jain, and Abbeel]{ho2020denoising}
Jonathan Ho, Ajay Jain, and Pieter Abbeel.
\newblock Denoising diffusion probabilistic models.
\newblock 2020.

\bibitem[Hu et~al.(2022)Hu, yelong shen, Wallis, Allen-Zhu, Li, Wang, Wang, and Chen]{hu2022lora}
Edward~J Hu, yelong shen, Phillip Wallis, Zeyuan Allen-Zhu, Yuanzhi Li, Shean Wang, Lu Wang, and Weizhu Chen.
\newblock Lo{RA}: Low-rank adaptation of large language models.
\newblock In \emph{International Conference on Learning Representations}, 2022.

\bibitem[Huang and Belongie(2017{\natexlab{a}})]{huang2017arbitrary}
Xun Huang and Serge Belongie.
\newblock Arbitrary style transfer in real-time with adaptive instance normalization.
\newblock In \emph{ICCV}, 2017{\natexlab{a}}.

\bibitem[Huang and Belongie(2017{\natexlab{b}})]{Huang2017ArbitraryST}
Xun Huang and Serge~J. Belongie.
\newblock Arbitrary style transfer in real-time with adaptive instance normalization.
\newblock \emph{2017 IEEE International Conference on Computer Vision (ICCV)}, 2017{\natexlab{b}}.

\bibitem[Jia et~al.(2023)Jia, Zhao, Chan, Li, Zhang, Gong, Hou, Wang, and Su]{jia2023taming}
Xuhui Jia, Yang Zhao, Kelvin C.~K. Chan, Yandong Li, Han Zhang, Boqing Gong, Tingbo Hou, Huisheng Wang, and Yu-Chuan Su.
\newblock Taming encoder for zero fine-tuning image customization with text-to-image diffusion models, 2023.

\bibitem[Kwon and Ye(2022)]{Kwon_2022_CVPR}
Gihyun Kwon and Jong~Chul Ye.
\newblock Clipstyler: Image style transfer with a single text condition.
\newblock In \emph{Proceedings of the IEEE/CVF Conference on Computer Vision and Pattern Recognition (CVPR)}, 2022.

\bibitem[Kwon and Ye(2023)]{kwon2023diffusionbased}
Gihyun Kwon and Jong~Chul Ye.
\newblock Diffusion-based image translation using disentangled style and content representation.
\newblock In \emph{The Eleventh International Conference on Learning Representations}, 2023.

\bibitem[Li et~al.(2023)Li, Li, Savarese, and Hoi]{li2023blip}
Junnan Li, Dongxu Li, Silvio Savarese, and Steven Hoi.
\newblock Blip-2: Bootstrapping language-image pre-training with frozen image encoders and large language models.
\newblock \emph{arXiv preprint arXiv:2301.12597}, 2023.

\bibitem[Liu and Deng(2015)]{7486599}
Shuying Liu and Weihong Deng.
\newblock Very deep convolutional neural network based image classification using small training sample size.
\newblock In \emph{2015 3rd IAPR Asian Conference on Pattern Recognition (ACPR)}, 2015.

\bibitem[Liu et~al.(2021)Liu, Lin, He, Li, Wang, Li, Sun, Li, and Ding]{Liu_2021_ICCV}
Songhua Liu, Tianwei Lin, Dongliang He, Fu Li, Meiling Wang, Xin Li, Zhengxing Sun, Qian Li, and Errui Ding.
\newblock Adaattn: Revisit attention mechanism in arbitrary neural style transfer.
\newblock In \emph{Proceedings of the IEEE/CVF International Conference on Computer Vision (ICCV)}, pages 6649--6658, 2021.

\bibitem[Loshchilov and Hutter(2019)]{loshchilov2018decoupled}
Ilya Loshchilov and Frank Hutter.
\newblock {Decoupled Weight Decay Regularization}.
\newblock In \emph{International Conference on Learning Representations}, 2019.

\bibitem[Meng et~al.(2022)Meng, Bau, Andonian, and Belinkov]{meng2022locating}
Kevin Meng, David Bau, Alex Andonian, and Yonatan Belinkov.
\newblock Locating and editing factual associations in {GPT}.
\newblock 2022.

\bibitem[Mou et~al.(2023)Mou, Wang, Xie, Zhang, Qi, Shan, and Qie]{mou2023t2i}
Chong Mou, Xintao Wang, Liangbin Xie, Jian Zhang, Zhongang Qi, Ying Shan, and Xiaohu Qie.
\newblock T2i-adapter: Learning adapters to dig out more controllable ability for text-to-image diffusion models.
\newblock \emph{arXiv preprint arXiv:2302.08453}, 2023.

\bibitem[Phillips and Mackintosh(2011)]{phillips:2011:wiki}
Fred Phillips and Brandy Mackintosh.
\newblock Wiki art gallery, inc.: A case for critical thinking.
\newblock \emph{Issues in Accounting Education}, 2011.

\bibitem[Radford et~al.(2021)Radford, Kim, Hallacy, Ramesh, Goh, Agarwal, Sastry, Askell, Mishkin, Clark, et~al.]{radford2021learning}
Alec Radford, Jong~Wook Kim, Chris Hallacy, Aditya Ramesh, Gabriel Goh, Sandhini Agarwal, Girish Sastry, Amanda Askell, Pamela Mishkin, Jack Clark, et~al.
\newblock {Learning Transferable Visual Models From Natural Language Supervision}.
\newblock In \emph{ICML}, 2021.

\bibitem[Ramesh et~al.(2021)Ramesh, Pavlov, Goh, Gray, Voss, Radford, Chen, and Sutskever]{pmlr-v139-ramesh21a}
Aditya Ramesh, Mikhail Pavlov, Gabriel Goh, Scott Gray, Chelsea Voss, Alec Radford, Mark Chen, and Ilya Sutskever.
\newblock Zero-shot text-to-image generation.
\newblock In \emph{Proceedings of the 38th International Conference on Machine Learning}, 2021.

\bibitem[Ramesh et~al.(2022)Ramesh, Dhariwal, Nichol, Chu, and Chen]{ramesh2022hierarchical}
Aditya Ramesh, Prafulla Dhariwal, Alex Nichol, Casey Chu, and Mark Chen.
\newblock Hierarchical text-conditional image generation with clip latents.
\newblock \emph{arXiv preprint arXiv:2204.06125}, 2022.

\bibitem[Razavi et~al.(2019)Razavi, van~den Oord, and Vinyals]{NEURIPS2019_5f8e2fa1}
Ali Razavi, Aaron van~den Oord, and Oriol Vinyals.
\newblock Generating diverse high-fidelity images with vq-vae-2.
\newblock In \emph{Advances in Neural Information Processing Systems}, 2019.

\bibitem[Rombach et~al.(2022)Rombach, Blattmann, Lorenz, Esser, and Ommer]{rombach2022high}
Robin Rombach, Andreas Blattmann, Dominik Lorenz, Patrick Esser, and Bj{\"o}rn Ommer.
\newblock High-resolution image synthesis with latent diffusion models.
\newblock In \emph{Proceedings of the IEEE/CVF conference on computer vision and pattern recognition}, 2022.

\bibitem[Ronneberger et~al.(2015)Ronneberger, Fischer, and Brox]{ronneberger2015u}
Olaf Ronneberger, Philipp Fischer, and Thomas Brox.
\newblock {U-Net: Convolutional Networks for Biomedical Image Segmentation}.
\newblock In \emph{MICCAI}, 2015.

\bibitem[Ruiz et~al.(2022)Ruiz, Li, Jampani, Pritch, Rubinstein, and Aberman]{ruiz2022dreambooth}
Nataniel Ruiz, Yuanzhen Li, Varun Jampani, Yael Pritch, Michael Rubinstein, and Kfir Aberman.
\newblock {DreamBooth: Fine Tuning Text-to-image Diffusion Models for Subject-Driven Generation}.
\newblock In \emph{CVPR}, 2022.

\bibitem[Ruiz et~al.(2023)Ruiz, Li, Jampani, Wei, Hou, Pritch, Wadhwa, Rubinstein, and Aberman]{ruiz2023hyperdreambooth}
Nataniel Ruiz, Yuanzhen Li, Varun Jampani, Wei Wei, Tingbo Hou, Yael Pritch, Neal Wadhwa, Michael Rubinstein, and Kfir Aberman.
\newblock Hyperdreambooth: Hypernetworks for fast personalization of text-to-image models, 2023.

\bibitem[Saharia et~al.(2022)Saharia, Chan, Saxena, Li, Whang, Denton, Ghasemipour, Gontijo~Lopes, Karagol~Ayan, Salimans, et~al.]{saharia2022photorealistic}
Chitwan Saharia, William Chan, Saurabh Saxena, Lala Li, Jay Whang, Emily~L Denton, Kamyar Ghasemipour, Raphael Gontijo~Lopes, Burcu Karagol~Ayan, Tim Salimans, et~al.
\newblock Photorealistic text-to-image diffusion models with deep language understanding.
\newblock 2022.

\bibitem[Schuhmann et~al.(2022)Schuhmann, Beaumont, Vencu, Gordon, Wightman, Cherti, Coombes, Katta, Mullis, Wortsman, Schramowski, Kundurthy, Crowson, Schmidt, Kaczmarczyk, and Jitsev]{schuhmann2022laionb}
Christoph Schuhmann, Romain Beaumont, Richard Vencu, Cade~W Gordon, Ross Wightman, Mehdi Cherti, Theo Coombes, Aarush Katta, Clayton Mullis, Mitchell Wortsman, Patrick Schramowski, Srivatsa~R Kundurthy, Katherine Crowson, Ludwig Schmidt, Robert Kaczmarczyk, and Jenia Jitsev.
\newblock {LAION}-5b: An open large-scale dataset for training next generation image-text models.
\newblock In \emph{Thirty-sixth Conference on Neural Information Processing Systems Datasets and Benchmarks Track}, 2022.

\bibitem[Shi et~al.(2023)Shi, Xiong, Lin, and Jung]{shi2023instantbooth}
Jing Shi, Wei Xiong, Zhe Lin, and Hyun~Joon Jung.
\newblock {InstantBooth: Personalized Text-to-Image Generation without Test-Time Finetuning}, 2023.

\bibitem[Song et~al.(2021)Song, Meng, and Ermon]{song2021denoising}
Jiaming Song, Chenlin Meng, and Stefano Ermon.
\newblock Denoising diffusion implicit models.
\newblock In \emph{International Conference on Learning Representations}, 2021.

\bibitem[Tewel et~al.(2023)Tewel, Gal, Chechik, and Atzmon]{tewel2023keylocked}
Yoad Tewel, Rinon Gal, Gal Chechik, and Yuval Atzmon.
\newblock Key-locked rank one editing for text-to-image personalization.
\newblock In \emph{ACM SIGGRAPH 2023 Conference Proceedings}, 2023.

\bibitem[Vaswani et~al.(2017)Vaswani, Shazeer, Parmar, Uszkoreit, Jones, Gomez, Kaiser, and Polosukhin]{NIPS2017_3f5ee243}
Ashish Vaswani, Noam Shazeer, Niki Parmar, Jakob Uszkoreit, Llion Jones, Aidan~N Gomez, \L~ukasz Kaiser, and Illia Polosukhin.
\newblock Attention is all you need.
\newblock In \emph{Advances in Neural Information Processing Systems}, 2017.

\bibitem[Wang et~al.(2023)Wang, Wang, Xie, Qi, Shan, Wang, and Luo]{wang2023styleadapter}
Zhouxia Wang, Xintao Wang, Liangbin Xie, Zhongang Qi, Ying Shan, Wenping Wang, and Ping Luo.
\newblock Styleadapter: A single-pass lora-free model for stylized image generation, 2023.

\bibitem[Wu et~al.(2021)Wu, Hu, Sheng, and Xu]{Wu_2021_ICCV}
Xiaolei Wu, Zhihao Hu, Lu Sheng, and Dong Xu.
\newblock Styleformer: Real-time arbitrary style transfer via parametric style composition.
\newblock In \emph{Proceedings of the IEEE/CVF International Conference on Computer Vision (ICCV)}, 2021.

\bibitem[Xiao et~al.(2023)Xiao, Yin, Freeman, Durand, and Han]{xiao2023fastcomposer}
Guangxuan Xiao, Tianwei Yin, William~T. Freeman, Frédo Durand, and Song Han.
\newblock Fastcomposer: Tuning-free multi-subject image generation with localized attention.
\newblock \emph{arXiv}, 2023.

\bibitem[Xu et~al.(2023)Xu, Guo, Wang, Huang, Essa, and Shi]{xu2023prompt}
Xingqian Xu, Jiayi Guo, Zhangyang Wang, Gao Huang, Irfan Essa, and Humphrey Shi.
\newblock Prompt-free diffusion: Taking" text" out of text-to-image diffusion models.
\newblock \emph{arXiv preprint arXiv:2305.16223}, 2023.

\bibitem[Ye et~al.(2023)Ye, Zhang, Liu, Han, and Yang]{ye2023ip-adapter}
Hu Ye, Jun Zhang, Sibo Liu, Xiao Han, and Wei Yang.
\newblock Ip-adapter: Text compatible image prompt adapter for text-to-image diffusion models.
\newblock \emph{arXiv preprint arxiv:2308.06721}, 2023.

\bibitem[Yu et~al.(2023)Yu, Wang, Zhao, Ghanem, and Zhang]{Yu_2023_ICCV}
Jiwen Yu, Yinhuai Wang, Chen Zhao, Bernard Ghanem, and Jian Zhang.
\newblock Freedom: Training-free energy-guided conditional diffusion model.
\newblock In \emph{Proceedings of the IEEE/CVF International Conference on Computer Vision (ICCV)}, 2023.

\bibitem[Zhang et~al.(2023{\natexlab{a}})Zhang, Rao, and Agrawala]{Zhang_2023_ICCV}
Lvmin Zhang, Anyi Rao, and Maneesh Agrawala.
\newblock Adding conditional control to text-to-image diffusion models.
\newblock In \emph{Proceedings of the IEEE/CVF International Conference on Computer Vision (ICCV)}, 2023{\natexlab{a}}.

\bibitem[Zhang et~al.(2022)Zhang, Tang, Dong, Huang, Ma, Lee, and Xu]{zhang2020cast}
Yuxin Zhang, Fan Tang, Weiming Dong, Haibin Huang, Chongyang Ma, Tong-Yee Lee, and Changsheng Xu.
\newblock Domain enhanced arbitrary image style transfer via contrastive learning.
\newblock In \emph{ACM SIGGRAPH}, 2022.

\bibitem[Zhang et~al.(2023{\natexlab{b}})Zhang, Huang, Tang, Huang, Ma, Dong, and Xu]{Zhang_2023_CVPR}
Yuxin Zhang, Nisha Huang, Fan Tang, Haibin Huang, Chongyang Ma, Weiming Dong, and Changsheng Xu.
\newblock Inversion-based style transfer with diffusion models.
\newblock In \emph{Proceedings of the IEEE/CVF Conference on Computer Vision and Pattern Recognition (CVPR)}, 2023{\natexlab{b}}.

\bibitem[Zhao et~al.(2023)Zhao, Chen, Chen, Bao, Hao, Yuan, and Wong]{zhao2023uni}
Shihao Zhao, Dongdong Chen, Yen-Chun Chen, Jianmin Bao, Shaozhe Hao, Lu Yuan, and Kwan-Yee~K Wong.
\newblock Uni-controlnet: All-in-one control to text-to-image diffusion models.
\newblock 2023.

\end{thebibliography}
}

\clearpage
\setcounter{section}{0}
\renewcommand{\thesection}{\Alph{section}}
\maketitlesupplementary

\section{Test Dataset}
\begin{figure}[t]
\begin{center}
   \includegraphics[width=1.\linewidth]{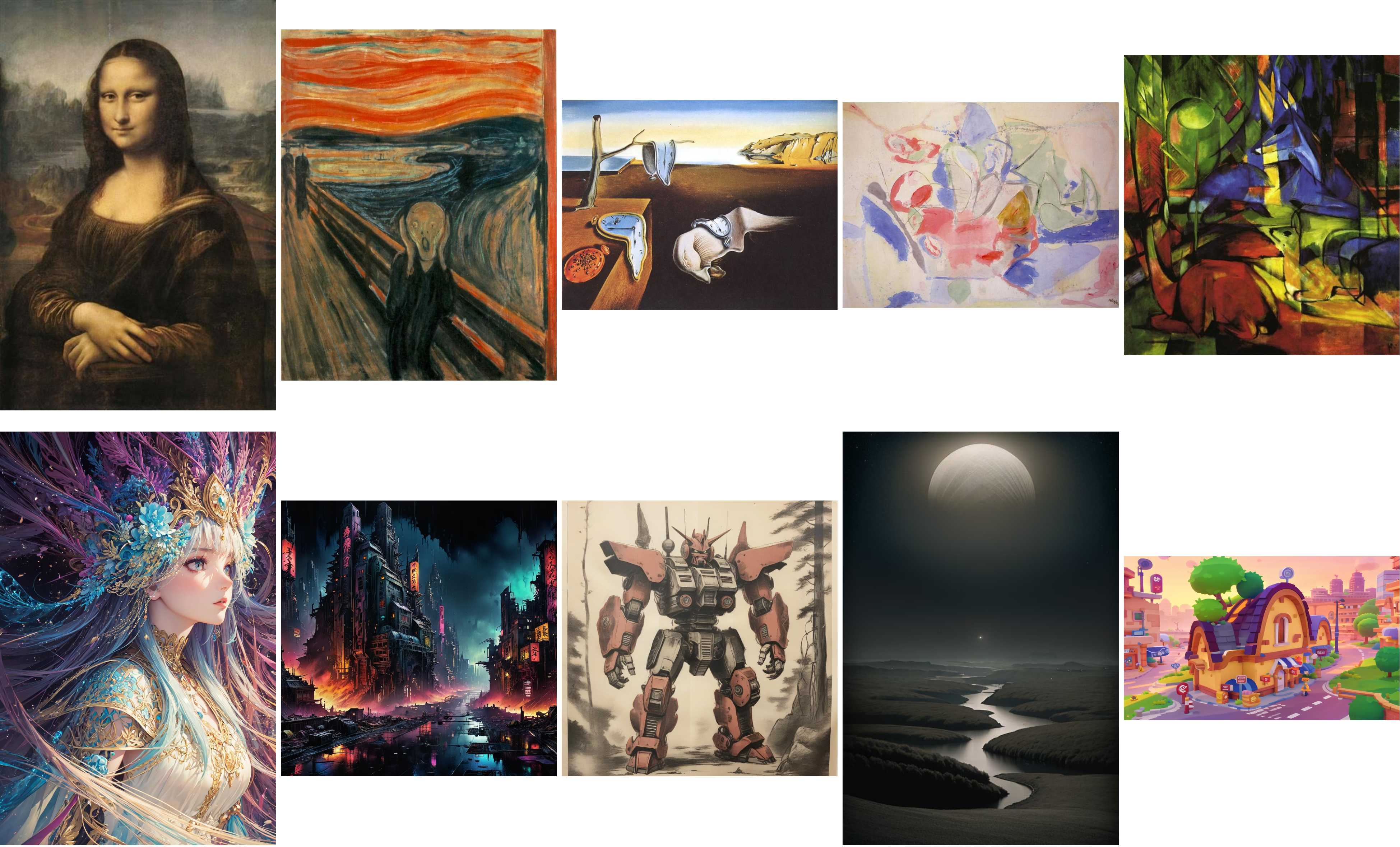}
\end{center}
   \caption{Style reference used in the test dataset for single-reference T2I style transfer.}
\label{fig:single_styles}
\end{figure}

\begin{figure}[t]
\begin{center}
   \includegraphics[width=1.\linewidth]{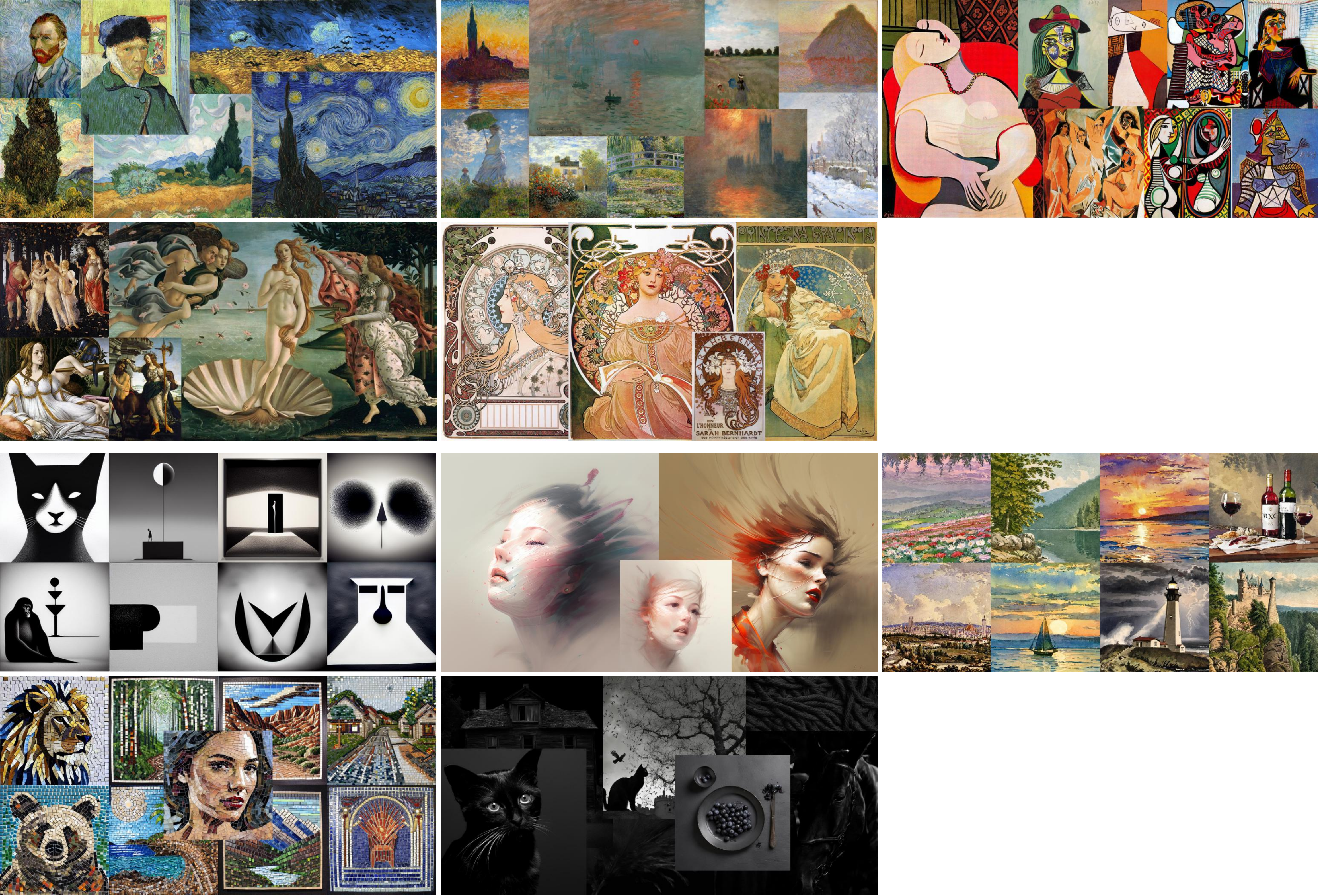}
\end{center}
   \caption{Collections of styles used in the test dataset for multi-reference T2I style transfer.}
\label{fig:multi_styles}
\end{figure}

We showcase the style references and prompts utilized in our quantitative evaluation and user study in Figures \ref{fig:single_styles} and \ref{fig:multi_styles}, and Table \ref{tab:prompt}.
Renowned artworks have been sourced from \url{https://www.wikiart.org/}, while others are synthesized images collected from \url{https://civitai.com/}.

\begin{table}[t]
    \centering
    \renewcommand{\arraystretch}{1.2}
    \vspace{0.05in}
    \resizebox{1.\linewidth}{!}{%
    \begin{tabular}{l|l}
      \toprule
      ``A robot'' & ``A curly-haired girl wearing a red dress, she is dancing''  \\
      ``A little boy playing football'' & ``A monkey playing with a banana''  \\
      ``A dog in the desert'' & ``A pine tree with a snowman hugging it''  \\
      ``A bird in a wood'' & ``A stone with a face carved on it, standing on a pedestal in a museum''  \\
      ``A kitten sleeping on a pillow'' & ``A modern house with a pool'' \\
      ``A daisy with a ladybug on it'' & ``A house covered with ice and snow'' \\
      ``Cherry blossom in full bloom'' & ``People in gondolas on the river'' \\
      ``solar system and galaxy'' & ``A woman sitting on grass near a tree'' \\
      ``A river with rapids and rocks'' & ``A windmill on a hillside'' \\
      ``Polar bears on the ice'' & ``A woman sitting on grass near a tree'' \\
      ``A snowy mountain peak'' & ``A castle with steps leading up to it'' \\
      ``A motorcycle'' & ``A couple sitting on a swing seeing aurora'' \\
      ``Sheep in autumn field'' & ``A woman wearing a baseball cap'' \\
      ``Volcano eruption'' & ``Eiffel Tower in heavy lightning storm'' \\
      ``A white rose'' & ``A house made of cardboard boxes'' \\
      ``A stone with a hole in it'' & ``A bronze sculpture of an old man'' \\
      ``A portrait of a rugby star'' & ``A colosseum near a waterfall'' \\
      ``A girl playing electric guitar'' & \\
      \bottomrule
    \end{tabular}
    }
    \caption{List of prompts used in the test dataset.}
    \label{tab:prompt}
\end{table}

\begin{figure}[t]
\begin{center}
   \includegraphics[width=1.\linewidth]{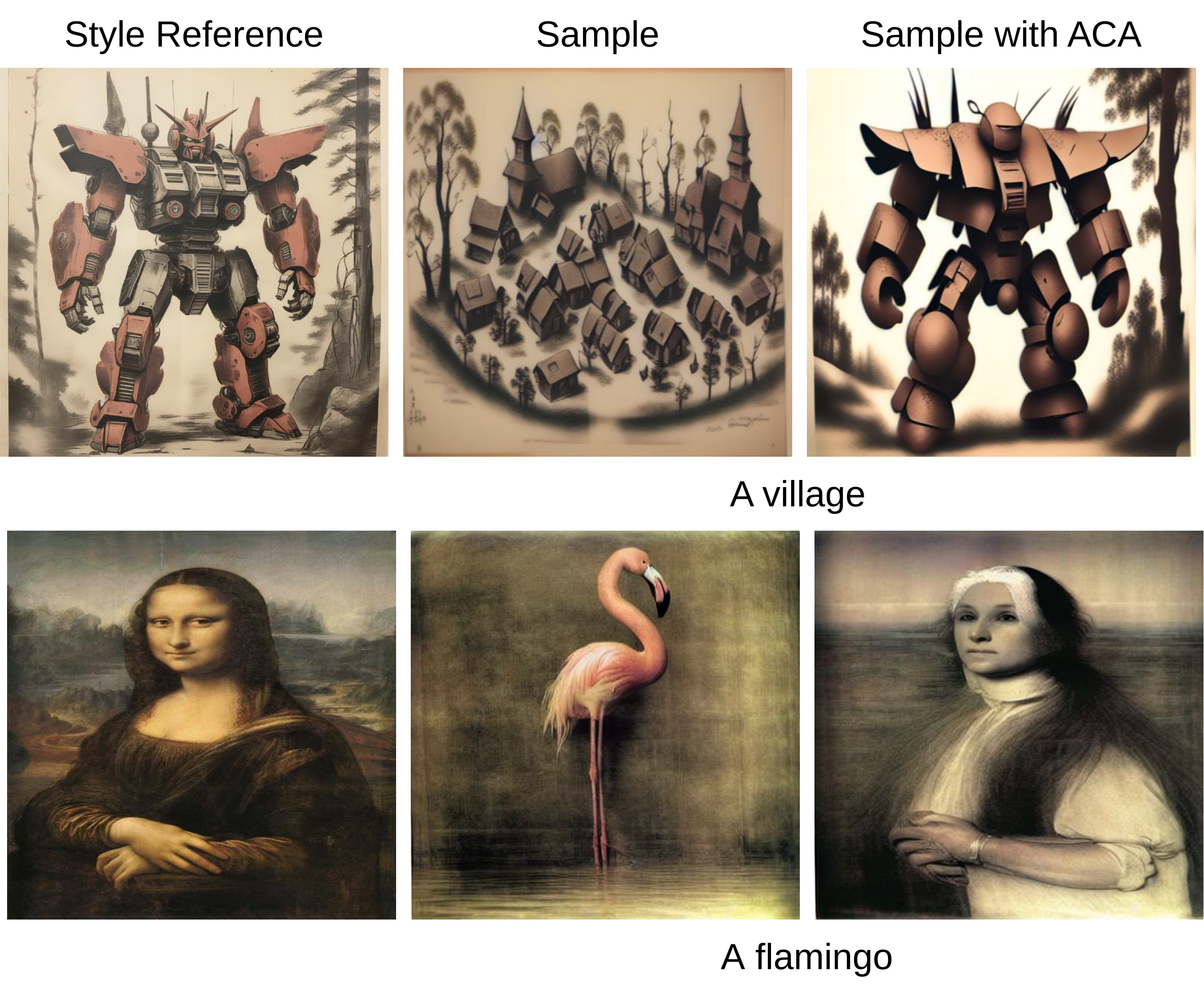}
\end{center}
   \caption{Ablative results of the ACA.}
\label{fig:abalation_aca}
\end{figure}

\begin{figure*}[t]
\begin{center}
   \includegraphics[width=1\linewidth]{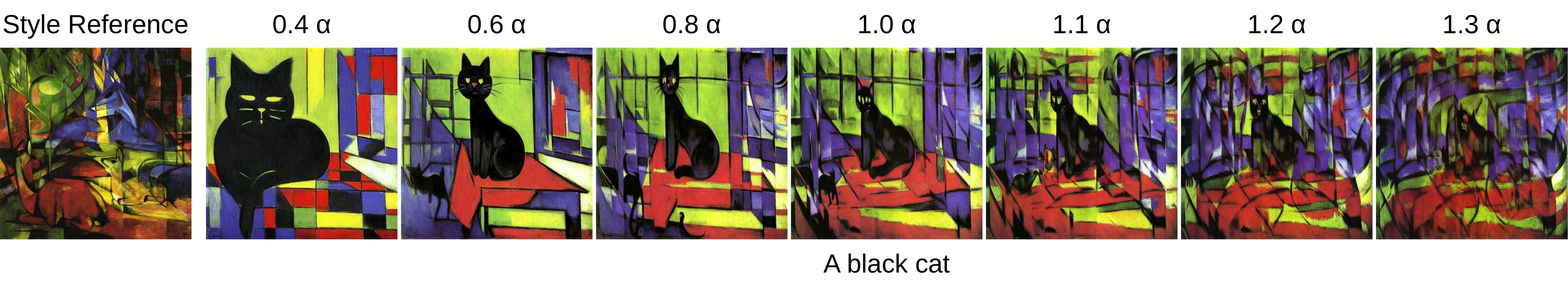}
\end{center}
   \caption{\textbf{Adapting $\alpha$.} ArtAdapter enables users to scaling the $\alpha$, effectively balance textual and style fidelity.}
\label{fig:alpha}
\end{figure*}

\begin{figure*}[t]
\begin{center}
   \includegraphics[width=1\linewidth]{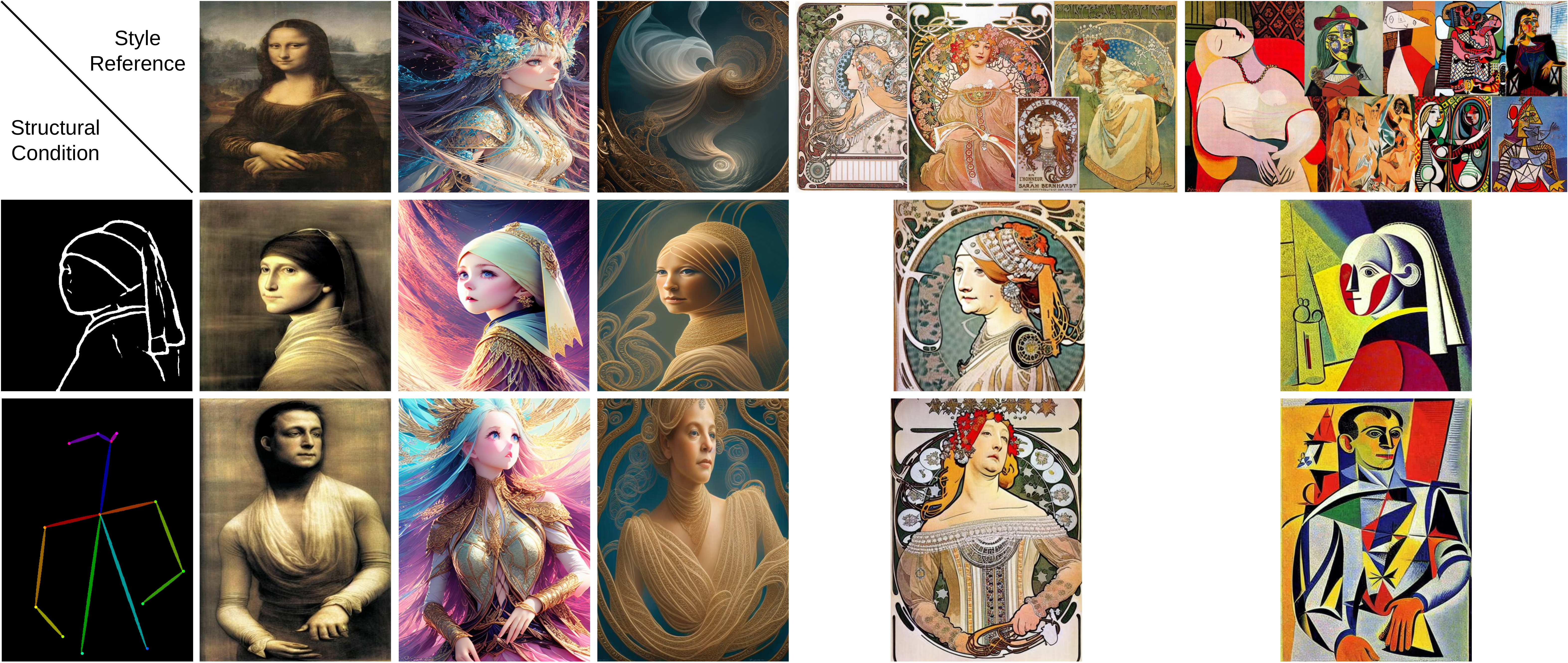}
\end{center}
   \caption{Results with additional structural controls.}
\label{fig:t2i_adapter}
\end{figure*}

\section{Details on User Study}
We randomly selected 40 generated images, encompassing the entire spectrum of multi-reference style collections and prompts.
Results from LoRA \cite{hu2022lora}, TI \cite{gal2023an}, and our ArtAdapter were presented in a randomized order for users to compare and rate on a seven-point scale.
Responses were solicited from 212 individuals, comprising both GenAI industry professionals and general users.
Each participant evaluated five image tuples randomly drawn from our selection, yielding an aggregate of 1060 assessments.

\section{Extended Ablation Study}

\subsection{Auxiliary Content Adapter}
Figure \ref{fig:abalation_aca} demonstrates how the Auxiliary Content Adapter (ACA) captures rough content structures, influencing the final results.
When employing ACA during the denoising process, the content structures from style references will be presented.
Consequently, by omitting ACA during inference, we can effectively remove these content structures while retaining the style representation.

\subsection{Adaptive $\alpha$}
\label{suppl:alpha}
Figure \ref{fig:alpha} illustrates the impact of the adaptive $\alpha$, as defined in Equation \ref{eq:explicit_adaptation}, during sampling.
By adjusting the scaling factor of $\alpha$, users can tailor the balance between the style elements from style references and the textual semantics.
A higher $\alpha$ scaling emphasizes style traits in the results, albeit at the expense of textual semantics alignment, while a lower scaling does the opposite.
This flexibility empowers users to generate results that align with their own preferences.

\section{More Qualitative Results}
\begin{figure*}[t]
\begin{center}
   \includegraphics[width=.9\linewidth]{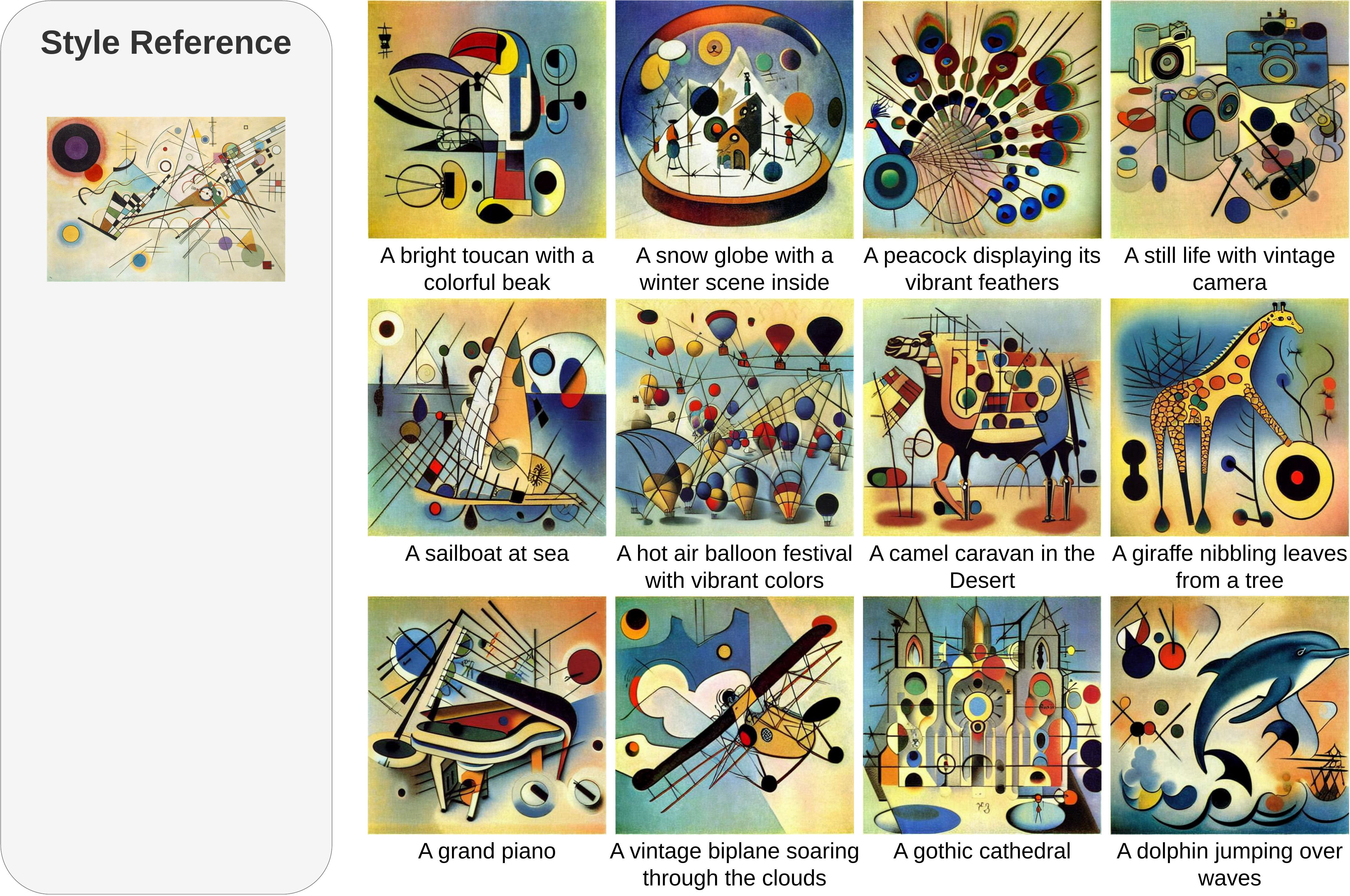}\\
   \vspace{5mm}
   \includegraphics[width=.9\linewidth]{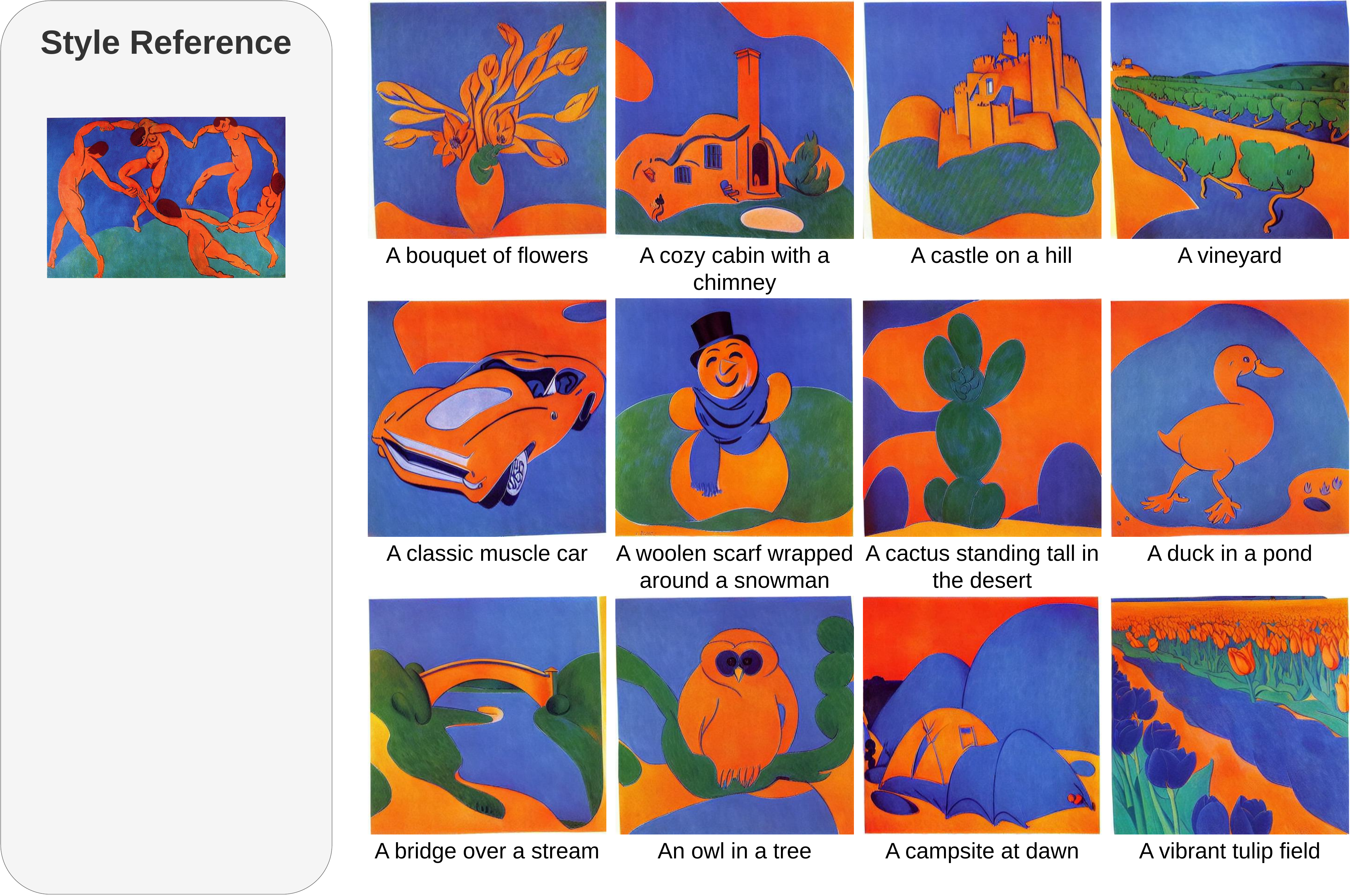}
\end{center}
   \caption{More results on single-reference T2I style transfer.}
\label{fig:more_single_1}
\end{figure*}

\begin{figure*}[t]
\begin{center}
   \includegraphics[width=.9\linewidth]{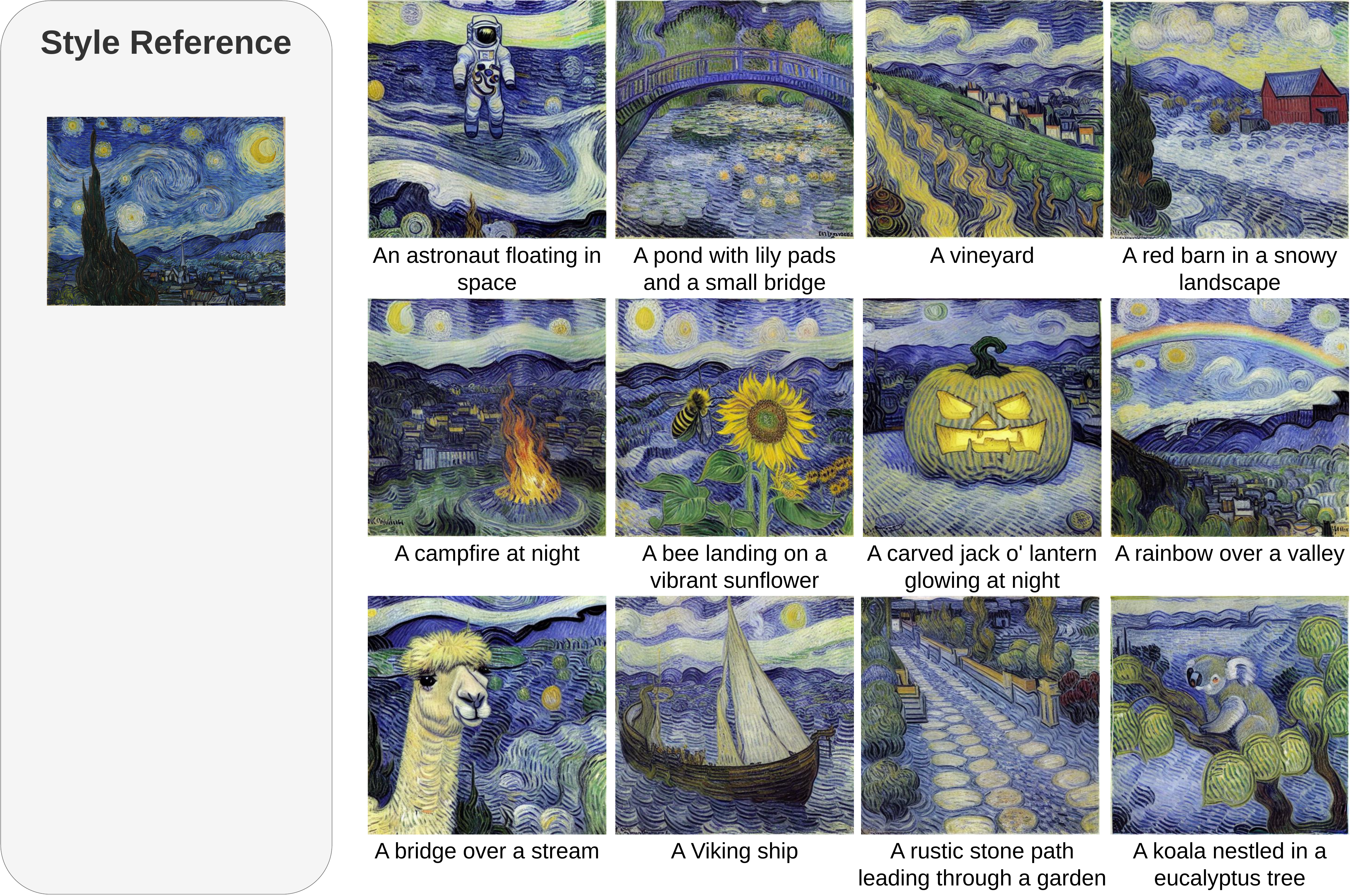}\\
   \vspace{5mm}
   \includegraphics[width=.9\linewidth]{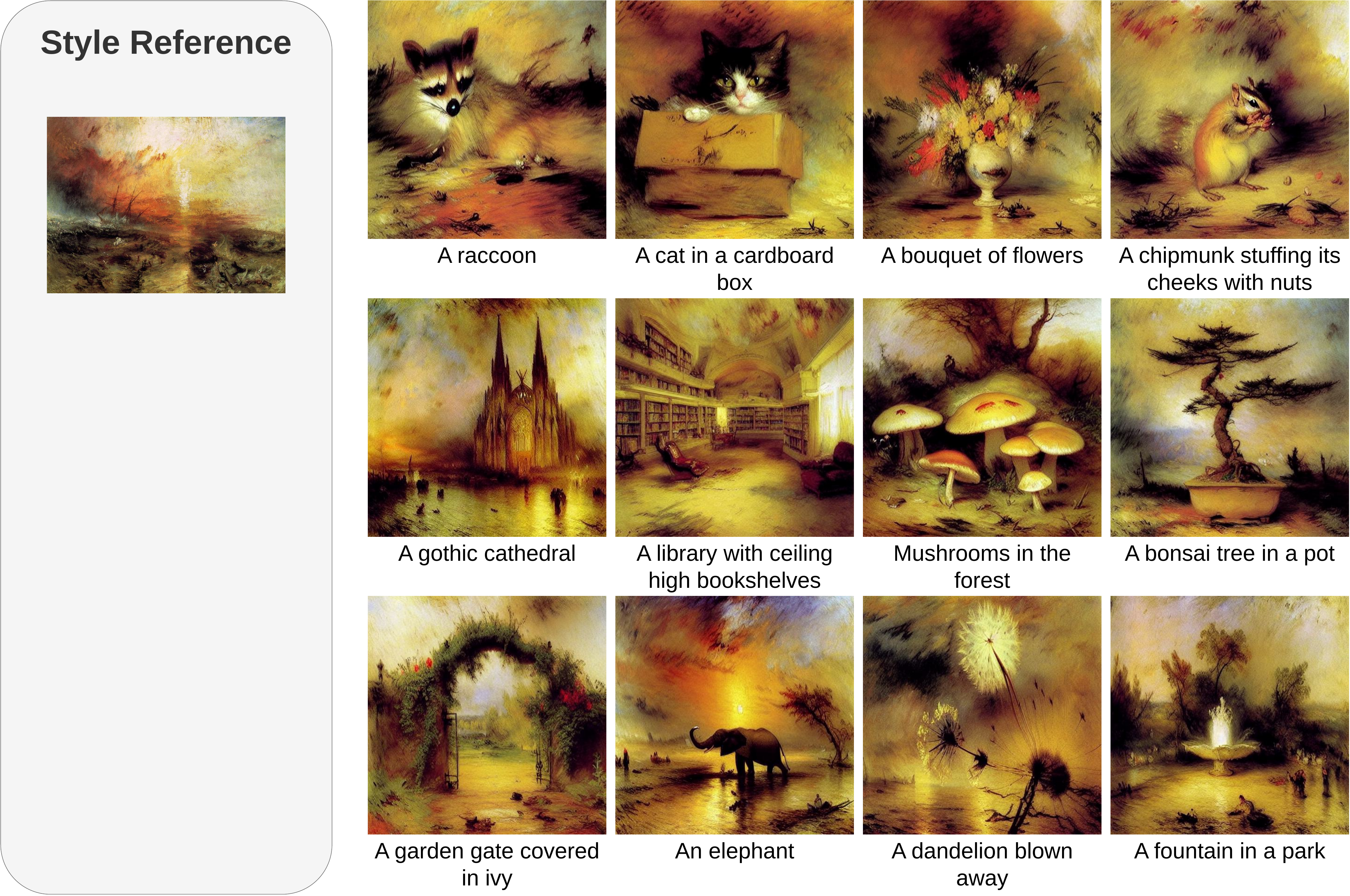}
\end{center}
   \caption{More results on single-reference T2I style transfer.}
\label{fig:more_single_2}
\end{figure*}

\begin{figure*}[t]
\begin{center}
   \includegraphics[width=.9\linewidth]{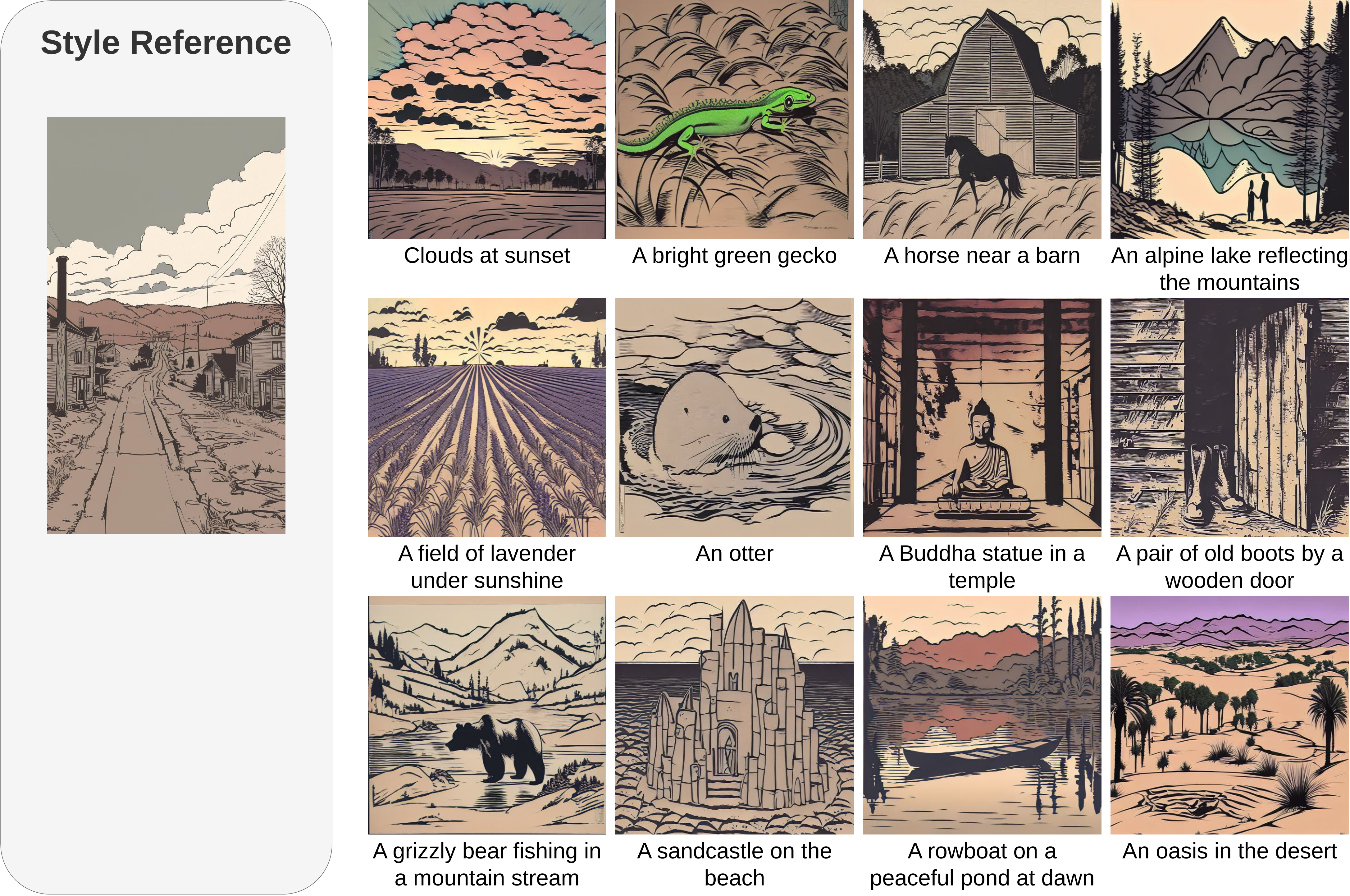}\\
   \vspace{5mm}
   \includegraphics[width=.9\linewidth]{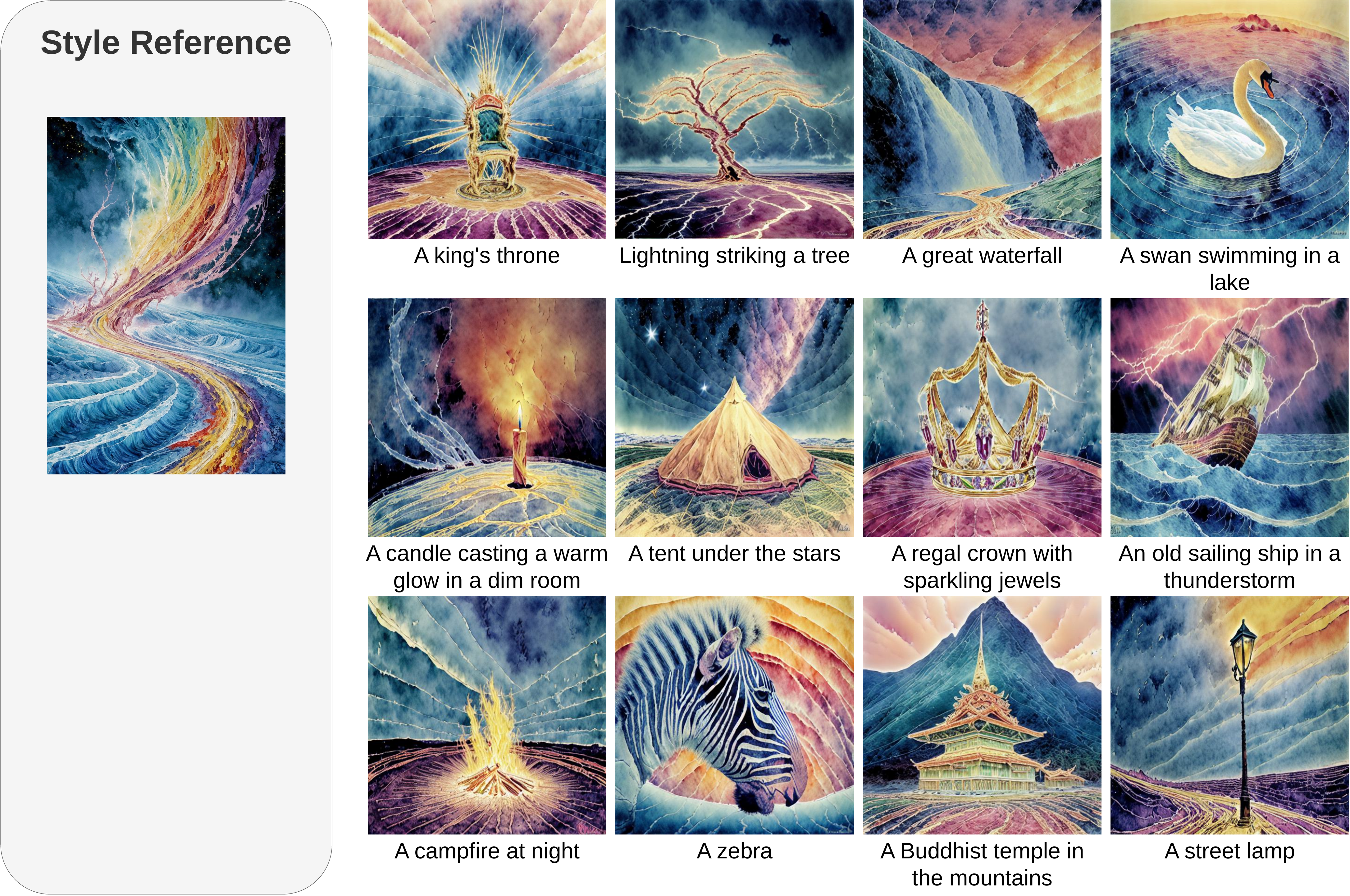}
\end{center}
   \caption{More results on single-reference T2I style transfer.}
\label{fig:more_single_3}
\end{figure*}

In Figure \ref{fig:t2i_adapter}, we demonstrate how effectively ArtAdapter integrates with the existing T2I-Adapter \cite{mou2023t2i}, showcasing its adaptability in exerting additional structural controls.
Further evaluations of ArtAdapter across a broader range of styles are presented in Figures \ref{fig:more_single_1}, \ref{fig:more_single_2}, and \ref{fig:more_single_3}.
These results underscore ArtAdapter's impressive performance across diverse styles.

\begin{figure*}[t]
\begin{center}
   \includegraphics[width=.9\linewidth]{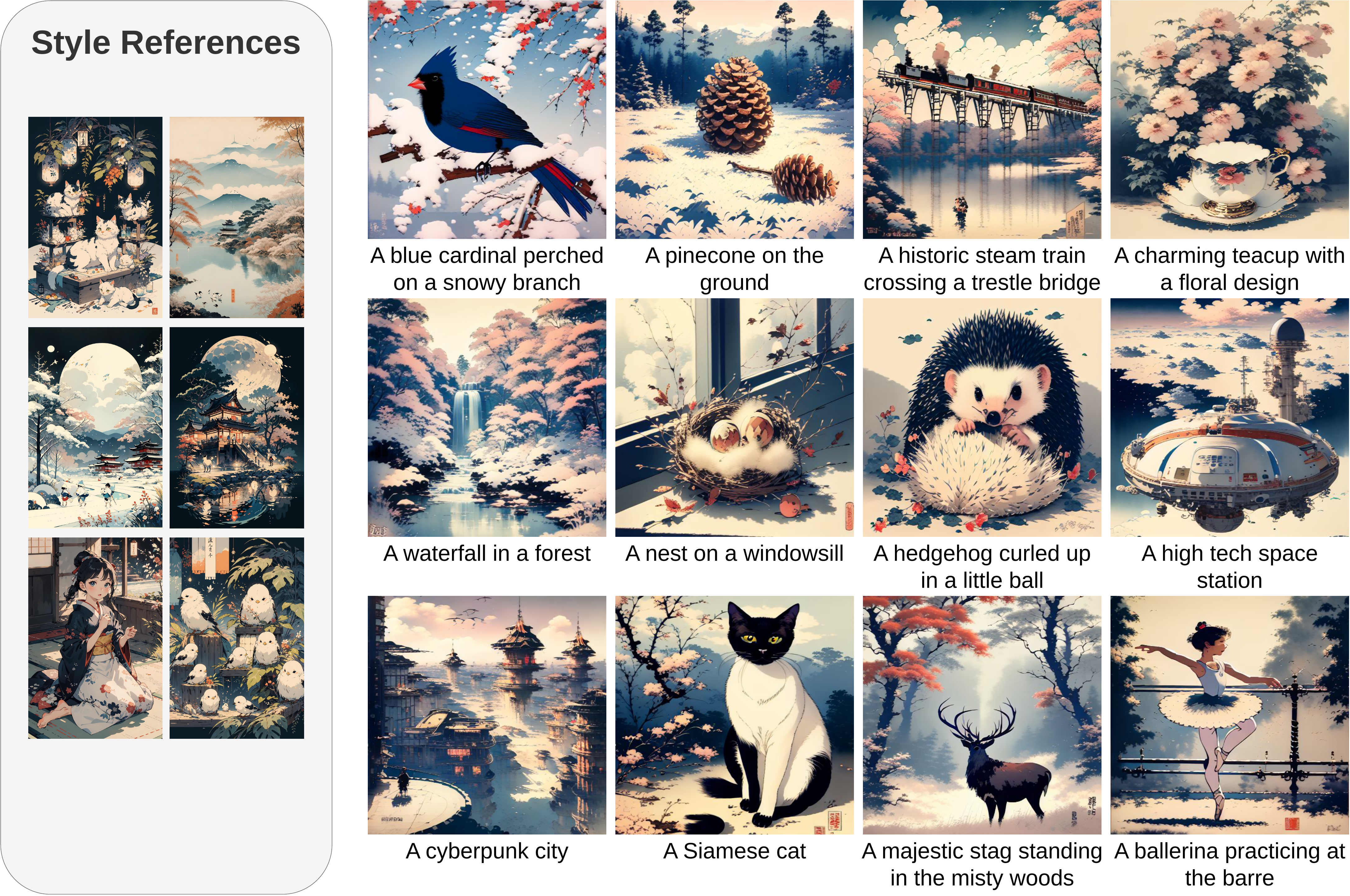}\\
   \vspace{5mm}
   \includegraphics[width=.9\linewidth]{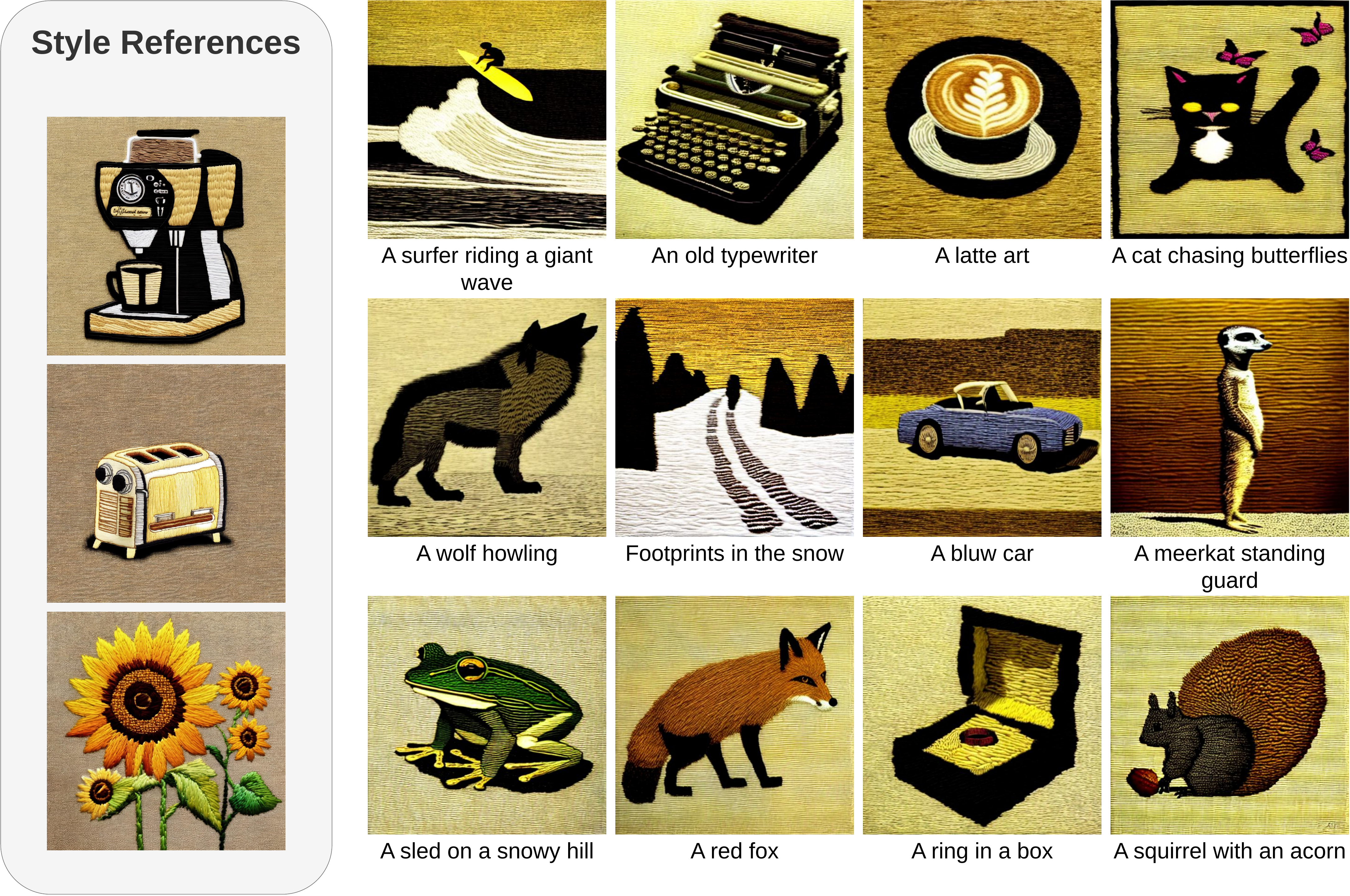}
\end{center}
   \caption{More results on multi-reference T2I style transfer.}
\label{fig:more_multi}
\end{figure*}


\end{document}